\documentclass[twocolumn]{svjour3}          % twocolumn
\smartqed  % flush right qed marks, e.g. at end of proof
\usepackage{graphicx}
\usepackage{amssymb}
\usepackage{latexsym}
\usepackage{epstopdf}
\usepackage{color}
\usepackage{soul}
\graphicspath{ {./Figures/}}
\usepackage{graphicx}
\usepackage{caption}
\usepackage{subcaption}
\usepackage[bookmarks=false]{hyperref}

\usepackage{amsmath}
\usepackage{algorithm,algorithmic}
\usepackage{paralist}
\usepackage{multirow}
\DeclareMathOperator*{\argmax}{arg\,max}
\usepackage{natbib}
\usepackage[inline]{enumitem}
\newcommand\sizz{1.1\columnwidth}% the size defined for showing the retinal patches 2
\newcommand\sizp{0.087\textwidth}% the size defined for showing the phantom patches
\usepackage{booktabs}
 \journalname{Biological Cybernetics}
\begin{document}
\setstcolor{red}

\title{Retrieving challenging vessel connections in retinal images by line co-occurrence statistics%\thanks{Grants or other notes
}
%\subtitle{}

\titlerunning{Line co-occurrences in retinal images}        % if too long for running head

\author{Samaneh Abbasi-Sureshjani \and 
	Jiong Zhang
%         \thanks{S. Abbasi-Sureshjani and M. Favali contributed equally to this work.}
	 \and  Remco Duits \and Bart ter Haar Romeny
	 %etc.
}

\authorrunning{S. Abbasi-Sureshjani} % if too long for running head

\institute{S. Abbasi-Sureshjani \and J. Zhang \and R. Duits \and B. ter Haar Romeny \at
              Department of Biomedical Engineering, Eindhoven University of Technology, P.O. Box 513, 5600 MB Eindhoven, The Netherlands \\
              \email{\{s.abbasi, J.Zhang1\}@tue.nl} 
              \and
              R. Duits \at 
              Department of Mathematics and Computer Science, Eindhoven University of Technology, P.O. Box 513, 5600 MB Eindhoven, The Netherlands \\
              \email{ r.duits@tue.nl}
               \and
              B. ter Haar Romeny  \at
			Department of Biomedical and Information Engineering,
			Northeastern University, 500 Zhihui Street, Shenyang,
			110167 China\\
			\email{bart.romeny@outlook.com}   		
}

\date{Received: date / Accepted: date}

\maketitle
%\tableofcontents

%----------Enter the sections
% !TEX root = EdgePaper.tex
\begin{abstract}
\sloppy Natural images contain often curvilinear structures, which might be disconnected, or partly occluded. Recovering the missing connection of disconnected structures is an open issue and needs appropriate geometric reasoning.
%{\color{blue}There is a close relation between the statistics of edges in natural images and the development of our perceptual system. The human brain is capable of grouping local edge elements to global contours using contextual information and their connections.} 
%In this work, we propose to find the line co-occurrences from retinal images, and use it as the cortical connectivity pattern. This statistical model is used to perform the perceptual grouping on retinal image patches automatically. To this end, 
We propose to find line co-occurrence statistics from the centerlines of blood vessels in retinal images and show its remarkable similarity to a well-known probabilistic model for the connectivity pattern in the primary visual cortex. Furthermore, the probabilistic model is trained from the data via statistics and used for automated grouping of interrupted vessels in a spectral clustering based approach. 
%To this end, an affinity matrix is created using this probabilistic model, followed by a self-tuning spectral clustering technique, which detects the salient blood vessels automatically.
%Here, an affinity matrix is created using the probabilistic model, followed by a self-tuning spectral clustering step to find the salient 
%%\st{perceptual units in the image, representing the} 
%blood vessels. 
Several challenging image patches are investigated around junction points, where successful results indicate the perfect match of the trained model to the profiles of blood vessels in retinal images. 
%strength of the trained model as indeed the connections among the blood vessels are detected successfully. 
Also, comparisons among several statistical models obtained from different datasets reveals their high similarity i.e., they are independent of the dataset. 
%Moreover, the comparisons of these kernels with the numerical models of the cortical connectivity confirm their close relations.
%Comparison of the statistical model to the probabilistic model
%%on the projective line bundle
%reveals a remarkably strong similarity. 
On top of that, the best approximation of the statistical model with the symmetrized extension of the probabilistic model on the projective line bundle is found with a least square error smaller than $2\%$.
Apparently, the direction process on the projective line bundle is a good continuation model for vessels in retinal images.
\keywords{Curvilinear structures \and Line co-occurrences \and Cortical connectivity \and Contextual affinity matrix \and Spectral clustering \and Perceptual grouping
}
\end{abstract}
% !TEX root = EdgePaper.tex
\section{Introduction}
\label{sec:intro}
\paragraph{Tracking curvilinear structures}
\sloppy Tree like structures such as the retinal vasculature, corneal nerve fibers, and roads from aerial photographs for cartography
%, and plant roots 
are widely studied both in quantitative computer-aided diagnosis systems in large-scale screening programs, and high-volume industrial settings. Delineation of curvilinear structures in these images is essential for investigating their characteristics. For instance, several studies highlighted the importance of using quantitative measurements of morphological and geometrical properties of blood vessels in retinal images for early diagnosis and prognosis of several diseases such as hypertension and diabetic retinopathy~\citep[e.g.][]{chapman2002peripheral,smith2004retinal}.
%{\color{red}~\citep[e.g.][]{chapman2002peripheral,smith2004retinal}}}.
%{chapman2002peripheral,smith2004retinal}
%\citep{chapman2002peripheral,foracchia2001extraction,smith2004retinal, fruttiger2007development}. 

\begin{figure}[htbp]
  \centering 
  		\begin{subfigure}{0.2\columnwidth} \includegraphics[width =1.1\columnwidth]{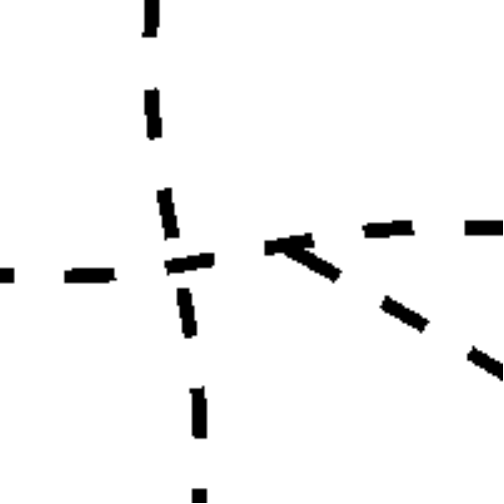}\caption{} \label{fig:phSeg}  \end{subfigure}
		\qquad
		\begin{subfigure}{0.2\columnwidth} \includegraphics[width =1.1\columnwidth]{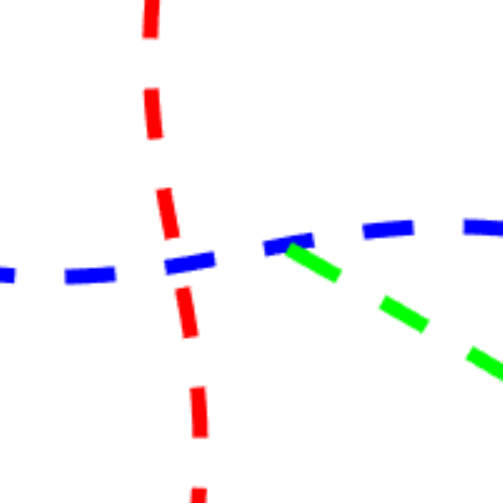}  \caption{}\label{fig:phGT} \end{subfigure}
	 \caption{(\subref{fig:phSeg}) A sample image with interrupted segments and (\subref{fig:phGT}) the salient units identified by our perceptual system}
  \label{fig:phantoms}
\end{figure}

Despite all improvements in the segmentation of curvilinear structures in two-dimensional images, the proposed methods often present limitations when two or more structures branch or cross, or when there are areas with missing information or interruptions~\citep{Fraz2012407}.
%\citep{Fraz2012407,sujatha2015connected,lin2009combining,chutatape1998retinal,xu2011vessel,gonzalez2010delineating}. 
Consequently, several tracking-based techniques provided solutions for preserving the connections in tree-shaped networks~\citep[e.g.][]{turetken2012automated,cheng2014tracing,Bekkers:2014aa,estrada2015tree,hu2015automated,de2016graph}. One of the common approaches has been to manually design cost functions, which penalized abrupt changes of the contextual features such as orientation, width and color. These costs were used in later stages in optimization or graph theory based techniques for constructing the full retinal vasculature network.
%Then either optimization techniques are used for finding the best network in the image with the minimum cost; or these costs are used in defining the weights of digraphs constructed based on an initial segmentation. The over-complete graphs are then narrowed down to several separated tree structures using graph-theory based techniques. 
In these methods, not only the cost functions were designed manually and depended on existing topological structures, but also tracing errors were often created due to the use of imperfect pixel-based vessel segmentations and skeletons that do not guarantee the connections among pixels belonging to one vessel.
%these methods often used segmentation and its skeleton as an initial step; therefore, their performances highly relied on the segmentation and its skeleton, and any errors in these two could propagate to later stages and cause tracing errors.
%Moreover, no solutions were proposed for situations that are more complex where some information is missing because of non-perfect imaging conditions, noise, interruptions, and occlusions.  
\paragraph{Geometry of primary visual cortex}
The human visual system is capable of interpreting visual scenes and of completing disconnected contours among interrupted segments, following the Gestalt law of good continuation~\citep{wertheimer1938laws}. 
Fig.~\ref{fig:phantoms} represents a sample interrupted phantom image (Fig.~\ref{fig:phSeg}) and the units detected by our perceptual system in different colors (Fig.~\ref{fig:phGT}).
Inspired by this capability 
% Considering this ability and being inspired by the solution proposed in several works~\citep{citti2006cortical,barbieri2014cortical,favali2015local,citti2014neuromathematics} in perceptual completion problems,
a new method was proposed by~\cite{favali2016analysis} for resolving the missing and complex connections among blood vessels at junction points in retinal images. 
%This method was examined only on retinal images, but it has the potential to be extended to a rich number of application areas including curvilinear structures. 
In this approach,
% after an initial segmentation and finding the vessel positions, 
first the image is lifted to the coupled space of positions and orientations using the so-called orientation score transformation introduced by~\cite{duits2007image}. The multi-orientation score is augmented with a contextual affinity matrix inspired by the long-range contextual connections between the multi-orientation pinwheel structures discovered in the primary visual cortex (V1)~\citep{hubel1962receptive,bosking1997orientation}. The affinities in this matrix are augmented by another similarity measurement of the feature of intensity and further processed in a spectral clustering step, which resulted in separate groups each representing one individual blood vessel.
%
%With a spectral clustering step the highest eigenvalues and their corresponding eigenvectors are determined. The dominant eigenvectors represent the individual blood vessels meeting at junction points. 
%The method is able to overcome the artifacts created during the segmentation, skeleton extraction and tracking steps as it is capable of grouping all the vessel pixels belonging to one vessel as a perceptual unit.

\sloppy{The cortical connectivity representing the contextual connections in V1 can be modeled as the fundamental solution of the time-independent Fokker-Planck (FP) equation for
Mumford's direction process~\citep{mumford1994elastica,williams1997stochastic,august2003sketches,citti2006cortical}.} Another closely related model for perceptual grouping of local orientations is a hypo-elliptic Brownian motion, whose Fokker-Planck equation describes hypo-elliptic diffusion without convection for the generator~\citep{citti2006cortical,duits2009line,agrachev2009intrinsic}.
These models explain well the notion of association fields introduced by~\cite{field1993contour} as a model for contour integration by the human visual system for image perception. %, following the Gestalt law of good continuation. %The connection between the neuro-geomterical model of the visual cortex and image completion was proved in~\citep{citti2006cortical} and implemented in~\citep{sanguinetti2008image} afterwards. 
There exist various numerical approximations, and exact solutions. %~\citep{duits2008explicit}.
For a recent overview and detailed comparison of all the solutions, see~\cite{zhang2014numerical} and references therein. Based on this study, the Fourier based technique
%~\citep{august2001curve} 
~\citep{duits2008explicit}
is the best approximation of the exact solution, having the smallest error both in the spatial and Fourier domains. The second best approximation is provided by the stochastic method based on the Monte Carlo simulation \citep{robert2013monte}. The stochastic solution was used by~\cite{favali2016analysis}. 

\paragraph{Edge statistics in natural images}
Second and higher order edge statistics are commonly used for representing the mutual relation between connected edges in natural images. Several studies~\citep[e.g.][]%{kruger1998collinearity,august2000curve,sigman2001common,geisler2001edge,geisler2008visual,elder2002ecological,sanguinetti2010model,perrinet2015edge}. 
{august2000curve,geisler2008visual,sanguinetti2010model,perrinet2015edge} investigated the statistics of the edges in our surrounding environment and their relation to the adaptation of connectivity patterns in our perceptual system.
%They confirmed the adaptation of connectivity patterns to the visual statistics. It is also discussed there, the direct contribution of these low level features obtained by neural activity in early visual areas to high-level judgments.
%They show that the contour grouping capability of the brain is directly related to the statistical co-occurrence of edge elements in natural images. 
These studies showed that the individual edges are dependent on each other, and the strongest characteristic determining their connections is their co-circularity relation. %Moreover, in several works the presence of parallel structures was reported as significant. 
%These statistics are proved to be useful for image processing applications. 
In the work by~\cite{august2000curve},
%, the curve indicator random field was introduced for defining the edge likelihood models used later for enhancement and completion purposes. They 
the edge statistics for several types of image categories were measured and it was shown that the resulting patterns were dependent on the structures available in the images. %So they were used for tuning their method for each image category individually. 
In a recent study by \cite{perrinet2015edge}, these low-level edge statistics were used for a high level judgmental task successfully. 
Moreover,~\cite{sanguinetti2010model} showed that there is a close relation between the statistics of edge co-occurrence and the probabilistic, geometric model of the cortical connectivity in V1. 
%They used a large set of natural images and by assuming the rotation-invariance property, they obtained a 3-dimensional statistical kernel. Then by comparing of this kernel with the numerical solutions of the forward and backward Fokker-Planck equations, the best match between these two kernels was found with the least square error of 2\%. 
%Their work helped in understanding the edge distribution in natural images and its connection to the geometry of V1 more clearly. It is also advantageous in tuning and improving the computer vision algorithms in contour completion problems. 
%In a recent work published in~\citep{perrinet2015edge}, the authors showed that these second-order edge statistics have the potential to be used for discriminating the animal images from man-made (non-animal) images. 
%Their results confirm the adaptation of connectivity patterns to the visual statistics. It is also discussed there, the direct contribution of these low level features obtained by neural activity in early visual areas to high-level judgments.
\paragraph{Our proposed method} 
We propose to train the probabilistic connectivity kernel using the line co-occurrence statistics on the lifted space of positions and ($\pi-$periodic) orientations ($\mathbb{R}^2\times P^1$) extracted directly from the retinal images. 
%In addition, we show the application of this trained model in retrieving the vessel connections at locations with complex structures in retinal images.
%This takes into account two facts: \begin{enumerate*}[label={\alph*)}] \item the relation between the contextual second-order line statistics in natural images and the geometrical model of cortical connectivity; and \item the dependency of these line statistics to existing structures in the images.\end{enumerate*} 
To this end, we make an adaptation of the direction process to the projective line bundles in $\mathbb{R}^2\times P^1$ instead of $\mathbb{R}^2\times S^1$ (the space of positions and $2\pi-$periodic orientations), as this extension is necessary for comparison of the probabilistic model to the statistical co-occurrences.
By comparing the statistical kernel to the symmetrized probability kernel, its best approximation resulting in the least square error is found. In fact, we show the relation between the probabilistic model of cortical connectivity and the edge statistics in our retinal imaging application is even closer when including both symmetrization and a projective line bundle structure.
 The dependency of the parameters of the statistical kernel with respect to the dataset is also investigated using different retinal image datasets, varying their resolutions and pixel sizes.

Finally, we show the application of this trained model in retrieving the vessel connections at locations with complex structures in retinal images. To this end an affinity matrix is created based on this statistical model and the similarities among vessel intensities and is analyzed in a self-tuning spectral clustering technique~\citep{zelnik2004self}, which does not need any parameter tuning and manual thresholding of the eigenvalues. It automatically determines the number of salient groups in the image by rotating the eigenvectors to create the maximally sparse representation and by minimizing a clustering cost defined accordingly.

%{\color{blue}The results of our analysis proves that Mumford direction process is a very good stochastic model for vessels. Besides, the symmetrized extension on the projective line bundle,  is remarkably close to the statistical kernels learned from the retinal images.}
%, supporting the symmetric line propagation model on the projective line bundle for completing partially corrupted blood vessels in retinal images.}

Summarizing, we demonstrate the following points in this article:
\begin{enumerate}[label=(\alph*)] % noitemsep,
\item  the statistical line co-occurrence kernel learned from retinal images matches remarkably well our symmetrized extension of the probabilistic model on the projective line bundle;
\item the statistical kernels do not change significantly over different datasets and are reproducible;
\item the low-level line statistics are successfully used to perform the high level task of grouping of the interrupted blood vessels in the retinal images automatically;
\item  Mumford's direction process is a very good stochastic model for connecting interrupted vessels in segmented retinal images.
\end{enumerate}

%The statistical kernel is later used for retrieving the vessel connections at locations with missing information or complex junctions. 
\paragraph{Paper structure}
The rest of the article is structured as follows. In Sect.~\ref{sec:theory}, the steps for deriving the line co-occurrences from retinal images and the theoretical details about modeling the cortical connectivity are described.
%, about the geometry of visual cortex, lifting the image to the space of positions and orientations and the cortical connectivity are explained. Besides, the steps for driving the edge co-occurrence statistics are described in detail. The cortically-inspired spectral clustering and its application in analysis of the retinal images are explained at the end of the section. 
In Sect.~\ref{sec:experiments}, after introducing the datasets, the resulting line co-occurrences are presented and compared against each other quantitatively and qualitatively. The best probability model approximating each kernel is presented afterwards. Application of the statistical kernel in retinal image analysis is presented at the end of the section. Finally, the results are discussed and the paper is concluded in Sect.~\ref{sec:conclusion}.

% !TEX root = EdgePaper.tex
\section{Methodology}
\label{sec:theory}
In this section, in addition to introducing the steps for extracting the line co-occurrences from retinal images, a numerical model of the connectivity kernel is proposed.
%explaining the steps used for driving the edge co-occurrences from retinal images, one of the numerical models of the cortical connectivity kernel is explained. Then the improving modifications for the vessel connectivity analysis are introduced. 
%{\color{magenta}In this section, we first explain how to lift the image to the roto-translation ($SE(2)$) group as the joint space of positions and orientations $\mathbb{R}^2\times S^1$, modelling the functional architecture of the primary visual cortex. Then the mathematical model of cortical connections and its existing solutions are described. The steps for extracting the edge co-occurrence from retinal images are explained afterwards. The application of the cortical connectivity kernel in analysing the connections in curvilinear images is explained at the end. }
%%------------------------------------------------------
\subsection{Line co-occurrence}
\label{sec:edgeStatistics}
In order to extract the line co-occurrences from retinal images, a similar approach as the method of~\cite{sanguinetti2010model} is used. 
%As explained in Sect.~\ref{sec:intro}, several studies were dedicated to extract the second order edge statistics from natural images. The setup used in this article for extracting the edge cross-correlations is very similar to the one proposed by~\citep{sanguinetti2010model}, with some differences. 
However, there are some differences. We only use retinal images, which include multiple elongated structures: the vessels; the vessel centerlines have been used instead of the edges (the resulting kernel is called \emph{line co-occurrences} rather than \emph{edge co-occurrences}) and no line polarity has been taken into account; the orientation score transformation has been used to find the orientation information at each point. 

%{\color{magenta}By defining $P^1 := S^1/\sim$, where $n_1 \sim n_2 \Leftrightarrow n_1 = \pm n_2$}, 
Recall that the projective circle $P^1$ is obtained from the normal circle $S^1$ by identifying antipodal points. The orientation score (OS) transform $\mathbb{R}^2\rightarrow\mathbb{R}^2\times P^1$ is obtained by correlating the input image $f$ with rotated isotropic (bi-directional) cake wavelets $\psi$~\citep{Bekkers:2014aa} in $n_\theta$ directions ($\theta\in[-\pi/2,\pi/2-\pi/n_\theta]$) as:
\begin{equation} 
 \begin{split}
U_f(\mathbf{x},\theta) & = ~(\overline{\mathbf{R}_\theta(\psi)} \star f)(\mathbf{x}) \\ 
& = \int_{\mathbb{R}^2} \overline{\psi(\mathbf{R}^{-1}_\theta (\mathbf{y}-\mathbf{x}))}f(\mathbf{y})\mathrm{d}\mathbf{y}
  \end{split} 
   \label{eq:OST} 
 \end{equation} 
where $\mathbf{R}_\theta$ is the 2D counter-clockwise rotation matrix, the overline denotes the complex conjugate and $\star$ denotes the correlation~\citep{duits2007image}.

The proposed method for finding the line statistics of the images of a dataset $S = \{I_1,I_2,\dots,I_n\}$, where $I_i\in \mathbb{R}^2$ is the $i^{th}$ retinal image, is explained step by step in Algorithm~\ref{alg:stepsHist}. The initial step is to create a set of interest vessel positions and orientations for each image. To obtain the vessel pixel locations the vessel ground truth is used, and the binary vessel centerlines ($I_{c,i},i=1,\dots,n$) are extracted in a standard morphological thinning approach~\citep{lam1992thinning}. So if a pixel at location $(x,y)$ belongs to a centerline, then $I_{c,i}(x,y)=1$, otherwise $I_{c,i}(x,y)=0$. If the ground truth is not available, then the vessel segmentation is obtained using one of the state-of-the-art techniques~\cite[e.g.][]{zhang2016robust}. The orientations at interest centerline positions are obtained by lifting the image using the OS transform (Step~\ref{st:OST1}), and finding the angles with the maximum response at these locations ($\theta_{m_i}(\mathbf{x})$ in Step~\ref{eq:MIPOS}). 
%{\color{blue}By defining $P^1 := S^1/\sim$, where $n_1 \sim n_2 \Leftrightarrow n_1 = \pm n_2$}, the OS transformation $\mathbb{R}^2\rightarrow\mathbb{R}^2\times P^1$ ($SE(2)$ group) is obtained by correlating the input image $f$ with rotated isotropic cake wavelets $\psi$~\citep{Bekkers:2014aa} in $n_\theta$ directions ($\theta\in[-\pi/2,\pi/2-\pi/n_\theta]$) as:
%\begin{equation} 
%% \begin{split}
%U_f(\mathbf{x},\theta)=(\overline{R_\theta(\psi)} \star f)(\mathbf{x})= \int_{\mathbb{R}^2} \overline{\psi(R^{-1}_\theta (\mathbf{y}-\mathbf{x}))}f(\mathbf{y})\mathrm{d}\mathbf{y}
%% \end{split}
% \label{eq:OST} 
% \end{equation} where $R_\theta$ is the 2D counter-clockwise rotation matrix, the overline denotes the complex conjugate and $\star$ denotes the correlation. 
It is worth mentioning that only the real part of the OS has been considered which acts as a ridge detector on the Gaussian profiles of blood vessels. Besides, the blood vessels in retinal images are darker compared to the background. As a result, they get negative responses (large absolute values) in this transformation. The negative sign used at Step \ref{eq:MIPOS} compensates for that. Using these locations and orientations, the set of interest points $L_i$ is created in Step~\ref{st:intSet} for each image.
%Finally, the isotropic cake wavelets in $n_\theta$ different directions ($\theta\in[-\pi/2,\pi/2]$) have been used in OS transformation.

\begin{figure}[htbp]
  \centering 
   \includegraphics[width = 0.8\columnwidth,trim={0 0.5in 0 0.5in},clip]{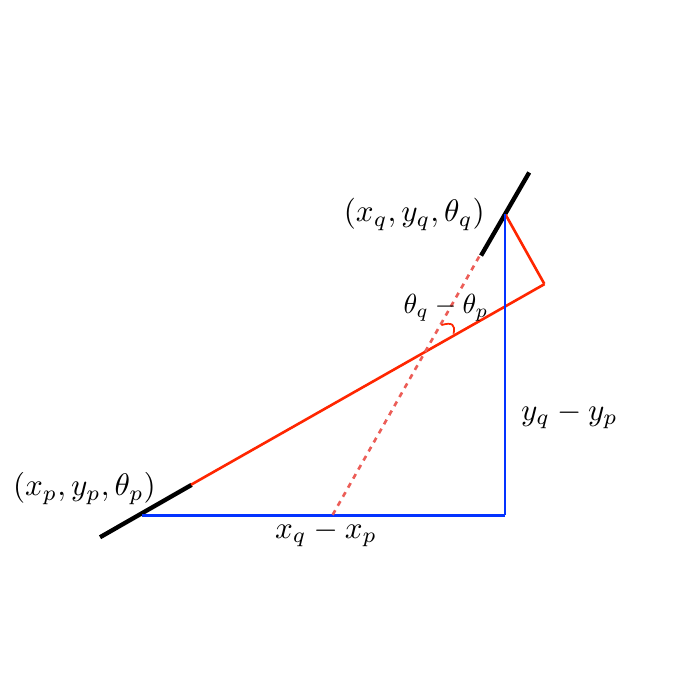}\caption{A sample pair of edges with positions and orientations of $(x_p,y_p,\theta_p)$ and $(x_q,y_q,\theta_q)$. The relative positions and orientations are depicted. Adapted from~\cite[Fig. 1]{sanguinetti2010model}.}\label{fig:edgeRelation}
\end{figure}

In the next step, pairs of interest points located at less than a certain distance ($d$) from each other are used to create a difference set $S_i^d$ considering the translation-invariance property (See Step~\ref{st:DifSet} and Fig.~\ref{fig:edgeRelation}). In order to make the set rotation-invariant, the relative positions are rotated with respect to the relative orientations and the shift-twist difference set $Q_i^d$ is created in Step~\ref{st:rotDifSet}.
%The set $Q_i^d$ can also be presented as:
%\begin{equation}
%Q_{i}^{d} =\{ h^{-1}g~|~ h=(\mathbf{x}_q,\theta_q),g=(\mathbf{x}_p,\theta_p)\}
%\end{equation}

%After preparing this set of positions and orientations per image, in Step~\ref{st:hist4D}, the number of pairs of edges located in a certain distance $d$ from each other (with relative positions of $|\Delta x|<d$ and $|\Delta y| <d$) and the orientations of $\theta_m$ and $\theta'_m$ are counted, having a translation-invariance assumption. This histogram is created per image and they are all accumulated in Step~\ref{st:accu}. Then in Step~\ref{st:3Dhist} with the rotation-invariance  assumption,
%the relative positions are rotated with the relative angles of $\Delta \theta = \theta'_m- \theta_m$ (see Fig~\ref{fig:edgeRelation}), and finally the rotation and translation invariant 3D histogram is obtained.  

By counting the number of iterations of relative positions and orientations in Step~\ref{st:3Dhisti} the statistical kernel per image is created. The statistical kernel of the entire dataset is obtained by accumulating the individual kernels of all the images in set $S$. Finally, the kernel is $l_1$-normalizedand and called the data-driven or statistical kernel (Step~\ref{st:final3Dhist}). The final dimension of this kernel is $(2d+1) \times (2d+1) \times n_\theta$. The parameter $d$ is selected heuristically during experiments.

In another approach, the artery/vein (AV) labels of vessels and the fact that arteries are not connected to veins are taken into account. By knowing these labels, the vessel profiles for arteries and veins are separated and their line co-occurrences are calculated individually ($K_{i,A}^{stat}$ and $K_{i,V}^{stat}$), using the similar steps as described before. Finally, the artery and vein histograms are added to each other to find the final histogram for each retinal image ($K_{i}^{stat}= K_{i,A}^{stat}+K_{i,V}^{stat}$). This is a more accurate assumption about the connections among vessel centerlines; however, it is only possible to use this setup if the AV labels are available. More details about datasets, parameter settings and  results are given in Sect.~\ref{sec:experiments}. 

\begin{algorithm*}
\caption{The proposed steps for obtaining the line co-occurrence statistics from a set $S$ of retinal images} 
	\begin{algorithmic} [1]
	\label{alg:stepsHist}
	%------------------------------------------------------
	\STATE Obtain the binary vessel centerlines ($I_{c,i}$) from the vessel ground truth of each image $I_i \in S, (i=1,\dots,n)$ by a standard morphological thinning approach.\label{st:centerline}
	%------------------------------------------------------
	\STATE Lift the original image ($I_i\in\mathbb{R}^2,\forall i=1,\dots,n$) to the rototranslation group ($U_{I_i}(\mathbf{x},\theta) \in \mathbb{R}^2\times P^1$) using isotropic cake wavelets rotated in $n_\theta$ directions in Eq.~\ref{eq:OST} so that $\theta \in \{-\pi/2,\dots,-\pi/2 + (\pi(n_\theta-1)/n_\theta)\}$.\label{st:OST1}
	%------------------------------------------------------
	\STATE Find the orientation with the highest real value of negative orientation score at each position $\mathbf{x}$ of $I_i,(i=1,\dots,n)$ as:
	$$\theta_{m_i}(\mathbf{x}) = \argmax_{\theta}Re(-U_{I_i}(\mathbf{x},\theta)) \label{eq:MIPOS}.$$
	%------------------------------------------------------
	\STATE Create a set of interest points for each image defined as $$L_i = \{(\mathbf{x},\theta_{m_i}(\mathbf{x}))~|~I_{c,i}(\mathbf{x}) = 1\},$$ where $\theta_{m_i}(\mathbf{x})$ is the dominant orientation at position $\mathbf{x}$.\label{st:intSet}
	%------------------------------------------------------
	\STATE Create a difference set defines as: $$S_i^d = \{(\mathbf{x}_p-\mathbf{x}_q,\theta_p-\theta_q)~|~(\mathbf{x}_p,\theta_p)\in L_i, (\mathbf{x}_q,\theta_q)\in L_i, \parallel\mathbf{x}_p - \mathbf{x}_q\parallel\leq d\}.$$ \label{st:DifSet}
	%------------------------------------------------------
	\STATE Create the shift-twist difference set as: $$Q_i^{d} = \{(\mathbf{R}_\theta^T(\mathbf{x}),\theta)~|~ (\mathbf{x},\theta)\in S_i^d\}.$$ \label{st:rotDifSet}
	%------------------------------------------------------
	\STATE Obtain $K_i^{stat}$ so that $$K_{i}^{stat}(\mathbf{x},\theta) =\text{the number of iterations of }(\mathbf{x},\theta)\in Q_i^{d}.$$ \label{st:3Dhisti}
	%------------------------------------------------------
	\STATE Calculate the total statistical kernel for the entire dataset as: $$K_{total}^{stat} = \sum_{i=1}^{n} K_i^{stat}$$ and normalize it as: $$K^{stat} = \frac{K_{total}^{stat}} {\parallel K_{total}^{stat} \parallel_{l_1}}.$$ \label{st:final3Dhist}
	\end{algorithmic}
\end{algorithm*}
\subsection{Cortical connectivity in $\mathbb{R}^2\times P^1$}
\begin{figure*}[htbp]
  \centering 
  		\begin{subfigure}{0.32\textwidth} \includegraphics[width =\columnwidth]{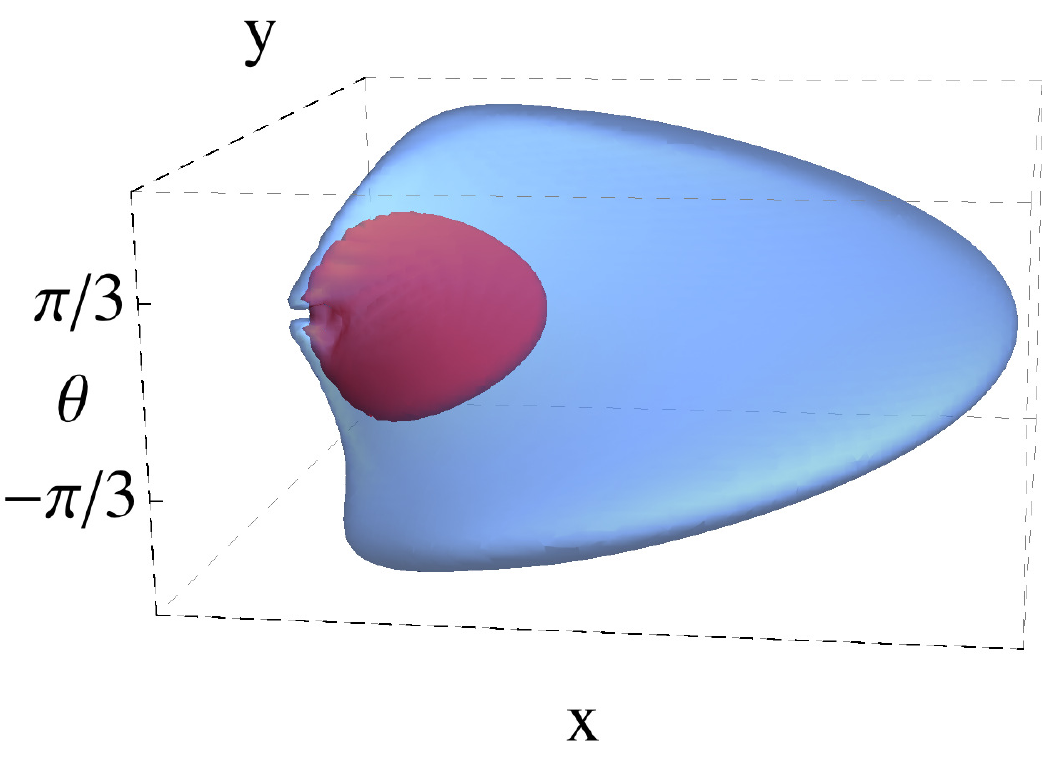}\caption{}\label{fig:forward2piView2}  \end{subfigure}\qquad
		\begin{subfigure}{0.32\textwidth} \includegraphics[width =\columnwidth]{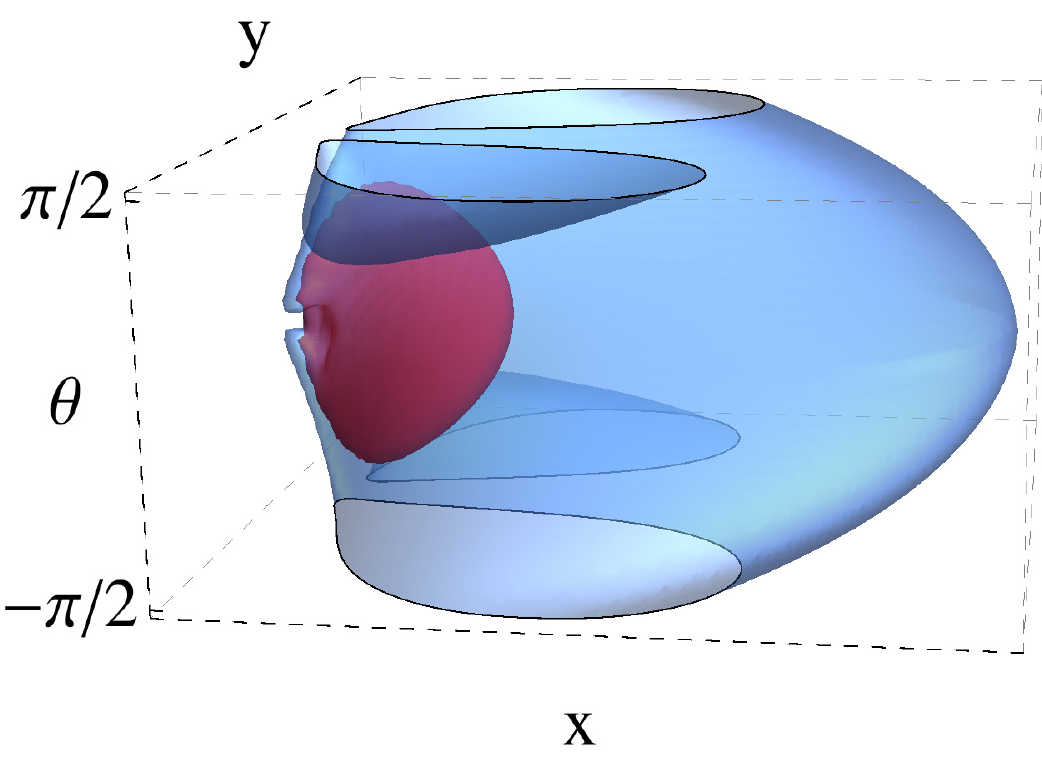}  \caption{}\label{fig:forwardpiView2} \end{subfigure}\qquad
		\begin{subfigure}{0.2\textwidth} \includegraphics[width =\columnwidth]{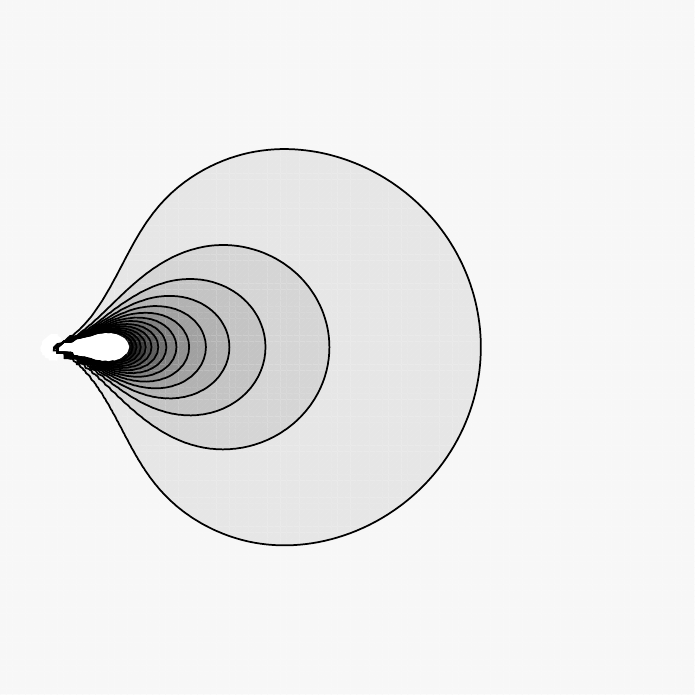}  \caption{}\label{fig:fPicontours} \end{subfigure}
		\begin{subfigure}{0.32\textwidth} \includegraphics[width =\columnwidth]{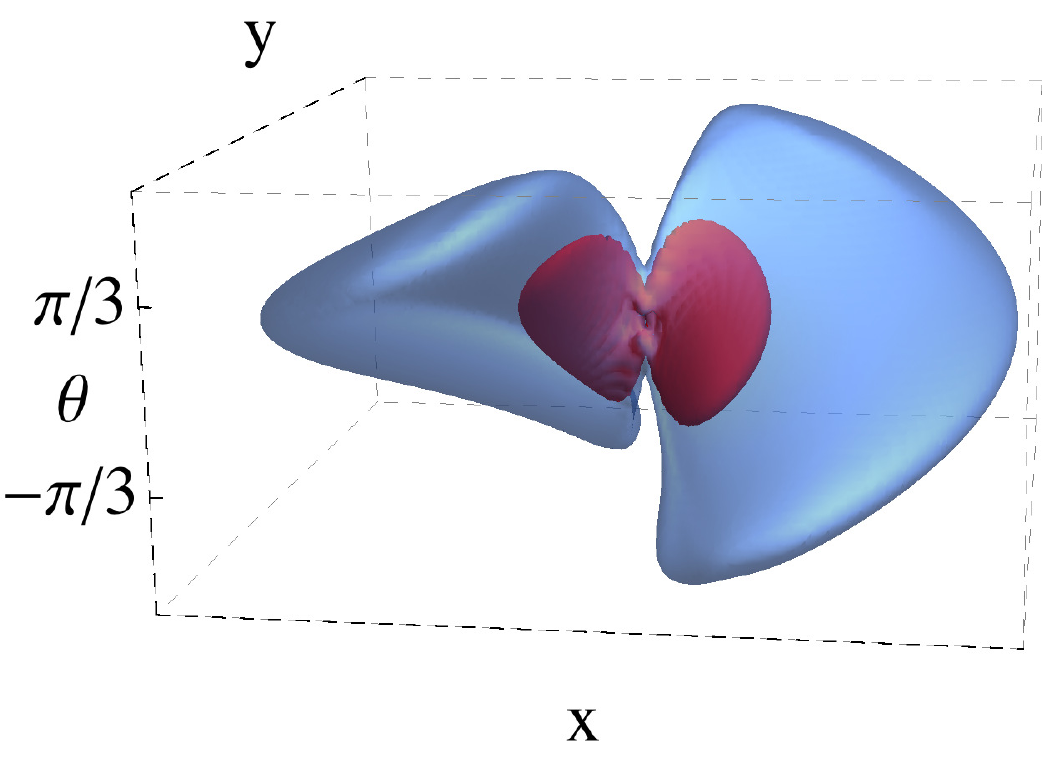}  \caption{}\label{fig:forback2pi} \end{subfigure}\qquad
		\begin{subfigure}{0.32\textwidth} \includegraphics[width =\columnwidth]{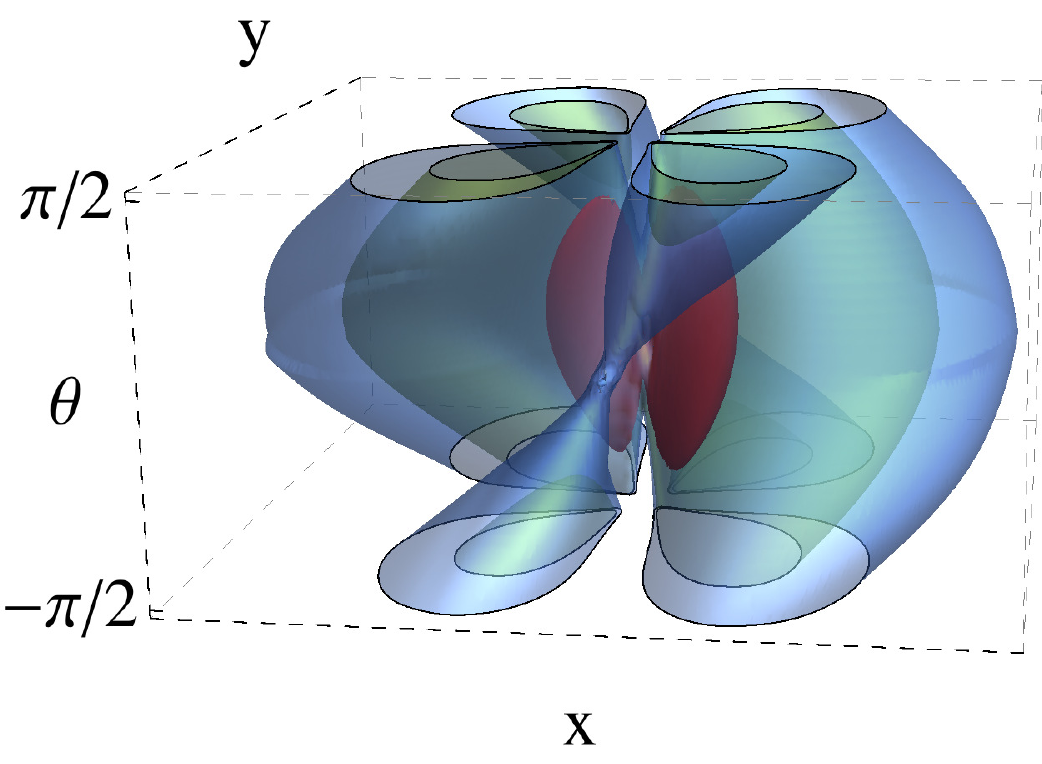}  \caption{}\label{fig:forbackpi} \end{subfigure}\qquad
		\begin{subfigure}{0.2\textwidth} \includegraphics[width =\columnwidth]{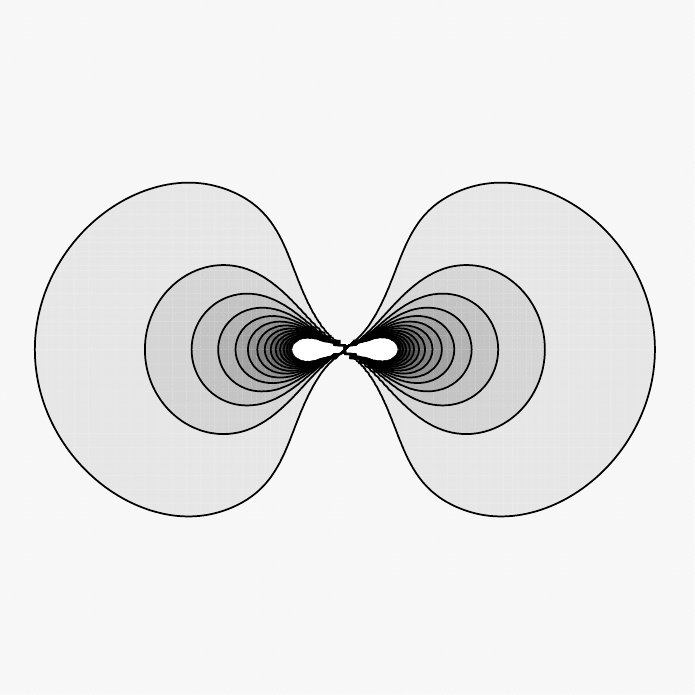}  \caption{}\label{fig:fbpicontours} \end{subfigure}
	 \caption{(\subref{fig:forward2piView2}) Forward kernel in $\mathbb{R}^2\times S^1$, (\subref{fig:forwardpiView2}) $\pi-$shifted forward kernel in $\mathbb{R}^2\times P^1$, (\subref{fig:fPicontours}) $xy$-marginal (obtained by integration over $\theta$) of the forward kernel,  (\subref{fig:forback2pi}) forward-backward kernel in $\mathbb{R}^2\times S^1$, (\subref{fig:forbackpi}) $k^{prob}$ in $\mathbb{R}^2\times P^1$, and (\subref{fig:fbpicontours}) $xy$-marginal of $k^{prob}$}
  \label{fig:fbkernels}
\end{figure*}

Considering Mumford's direction process in the differential structure of the sub-Riemannian SE(2) group, the fundamental solution of the FP equation represents the probability of having a contour at a certain position and orientation, starting from a reference position and orientation. In order to model the cortical connectivity kernel in $\mathbb{R}^2\times P^1$ (projective line bundle), we propose to create the connectivity kernel by adding the solutions of the FP equation in forward and backward directions in $\mathbb{R}^2\times S^1$ (for symmetrization) and the $\pi-$shifted solutions (for taking into account the final periodicity). Therefore, the connection probability between two points in $\mathbb{R}^2\times P^1$ is obtained by:
\begin{equation}
\begin{split}
k^{prob}((\mathbf{x},\theta),(\mathbf{x'},\theta')) =  \\
\frac{1}{4} \bigg( 
\Gamma \Big( (\mathbf{x},\theta),(\mathbf{x'},\theta') \Big )  +  
 \Gamma\Big((\mathbf{x'},\theta'),(\mathbf{x},\theta) \Big)   +  \\  
  \Gamma\Big((\mathbf{x},\theta+\pi),(\mathbf{x'},\theta')\Big) +  
  \Gamma\Big((\mathbf{x'},\theta'+\pi),(\mathbf{x},\theta)\Big)  \bigg) 
  \end{split}
\label{eq:2PitoPi}
\end{equation}
where $\Gamma$ is the fundamental solution of the time integrated FP equation centered around $(\mathbf{x'},\theta')$ represented as:
\begin{equation}
\Gamma \Big( (\mathbf{x},\theta),(\mathbf{x'},\theta') \Big ) = R^\mathbf{D}_\alpha(\mathbf{R}^T_\theta(\mathbf{x-x'}),\theta-\theta')
\label{eq:GammaSolution}
\end{equation}
with the resolvent kernel $R^\mathbf{D}_\alpha$ obtained by integrating Green's function $K_t^\mathbf{D}: SE(2) \rightarrow \mathbb{R}^+$ as:
\begin{equation}
R^\mathbf{D}_\alpha(\mathbf{x},\theta) = \alpha\int_0^\infty K_t^\mathbf{D}(\mathbf{x},\theta)\mathrm{e}^{-\alpha t}\mathrm{d}t
\label{eq:resolvKernel}
\end{equation}
which is the solution of the following PDE:
\begin{equation}
(\cos\theta\partial_x + \sin\theta\partial_y-\mathbf{D}\partial_\theta^2-\alpha I) R^\mathbf{D}_\alpha= \alpha \delta_e.
\label{eq:pde}
\end{equation}
Note that this PDE is defined on $\mathbb{R}^2\times S^1$ and not on $\mathbb{R}^2\times P^1$ as the first order part flips when applying $\theta\rightarrow\theta+\pi$. Therefore, in Eq.~\ref{eq:2PitoPi}, besides a $\pi-$shift, we need inversion invariance yielding a double symmetric kernel (see Fig.~\ref{fig:fbkernels}).
%------
%------

In this work the numerical solution has been created using the Fourier-based technique~\citep{duits2008explicit}, because it is not only the best approximation to the exact solution, but also computationally the least expensive one compared to the other solutions~\citep{zhang2014numerical}. Referring to \cite[Fig.13][]{zhang2014numerical}, the slight spatial blurring with $0<s\ll1$ corresponds to a one-pixel bin used in statistical kernel. So the exact probability  kernel (with small $s$)  is considered with  the same resolution as the statistical kernel.  
The final probability kernel and the statistical data-driven kernels are compared in the spatial domain. The key parameters in creating the probability kernel are $\alpha$ and $D_{33}=\sigma^2/2$, which determine the expected life time of the resolvent kernel ($E(T)=1/\alpha$) and the diffusion matrix ($D = \textrm{diag}\{0,0,D_{33}\}$ ) respectively (see~\cite{zhang2014numerical} for more details). To have uniform notations in the rest of the article, we assume the following relations hold for both probabilistic and statistical kernels:
\begin{equation}
\begin{split}
k^{stat} (h, g) =k^{stat}( e, h^{-1}g) = K^{stat}(h^{-1}g)\\
k^{prob}(h,g)=k^{prob}(e,h^{-1}g)=K^{prob}(h^{-1}g)\\
\end{split}
\end{equation}
where $g,h\in L_i$ for $i=1,\dots,n$. 

As a side note, the exact non-symmetrized kernel on $\mathbb{R}^2 \rtimes P^1$ is given by:
\begin{equation}
R_{\alpha}^{\mathbb{R}^{2} \rtimes P^{1}}(\mathbf{x},\theta)= \big{(}\mathcal{F}^{-1}_{\mathbb{R}^2}[ \mathbf{\omega} \mapsto \sum \limits_{n \in \mathbb{Z}} \hat{R}_\alpha^{\mathbf{D},\mathbf{a},\infty}(\mathbf{\omega},\theta + n \pi)] \big{)}(\mathbf{x})
\end{equation}
where $\hat{R}_\alpha^{\mathbf{D},\mathbf{a},\infty}$ is expressed in two Mathieu functions in~\cite[Eq.5.5,][]{zhang2014numerical}. Alternatively, one may take the simplified exact solution on $\mathbb{R}^2\times S^1$ \cite[Eq. 5.11][]{zhang2014numerical} and consider only the first and third terms in Eq.~\ref{eq:2PitoPi}.

\section{Experiments}
\label{sec:experiments}
In this section, first the datasets and the parameters used for finding the data-driven kernels are introduced. Then the results of the comparison of the data-driven kernels against each other, and comparison of these statistical kernels with the probability kernels are presented. At the end, the application of the data-driven kernel in retrieving vessel connections is explained. 
%-----------------------------------------------------------
\subsection{Material}
\begin{figure*}[htbp]
  \centering 
 	 \begin{subfigure}[b]{0.2\textwidth}
	 	 \includegraphics[width=0.95\columnwidth]{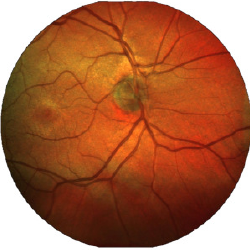}
	  	\includegraphics[width =\columnwidth]{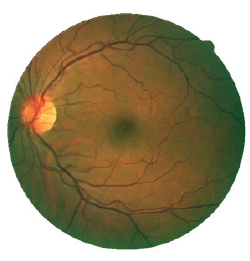} 
	   	 \caption{Original} \label{fig:FullRetinaOrig}
	\end{subfigure}
	\hspace{-0.1in}
	\begin{subfigure}[b]{0.2\textwidth}  
	 	\includegraphics[width =0.95 \columnwidth]{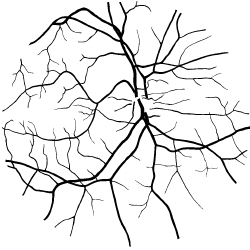}  
		\includegraphics[width = \columnwidth]{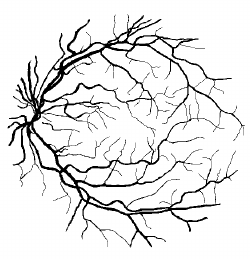}
	 	\caption{Vessel labels} \label{fig:FullRetinaGT}
	 \end{subfigure}
	 \hspace{-0.1in}
	\begin{subfigure}[b]{0.2\textwidth}
		 \includegraphics[width = 0.95\columnwidth]{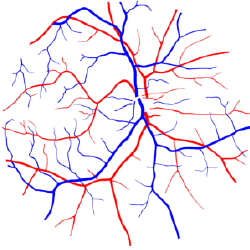} 
		  \includegraphics[width = \columnwidth]{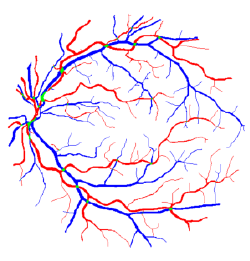}
		\caption{AV labels}  \label{fig:FullRetinaAV}
	\end{subfigure}
	\hspace{-0.1in}
	\begin{subfigure}[b]{0.2\textwidth} 
		\includegraphics[width =0.95\columnwidth]{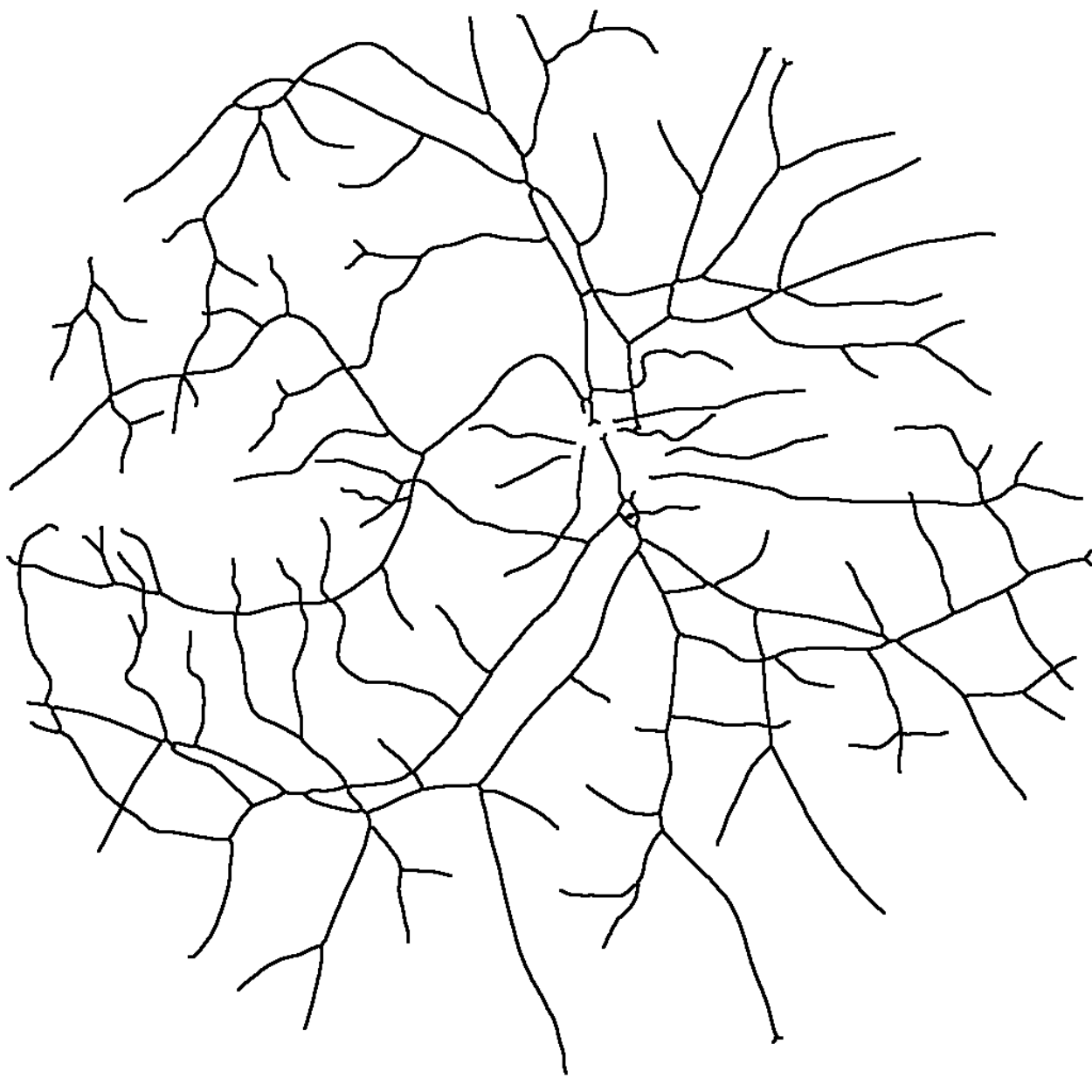} 
		\includegraphics[width =\columnwidth]{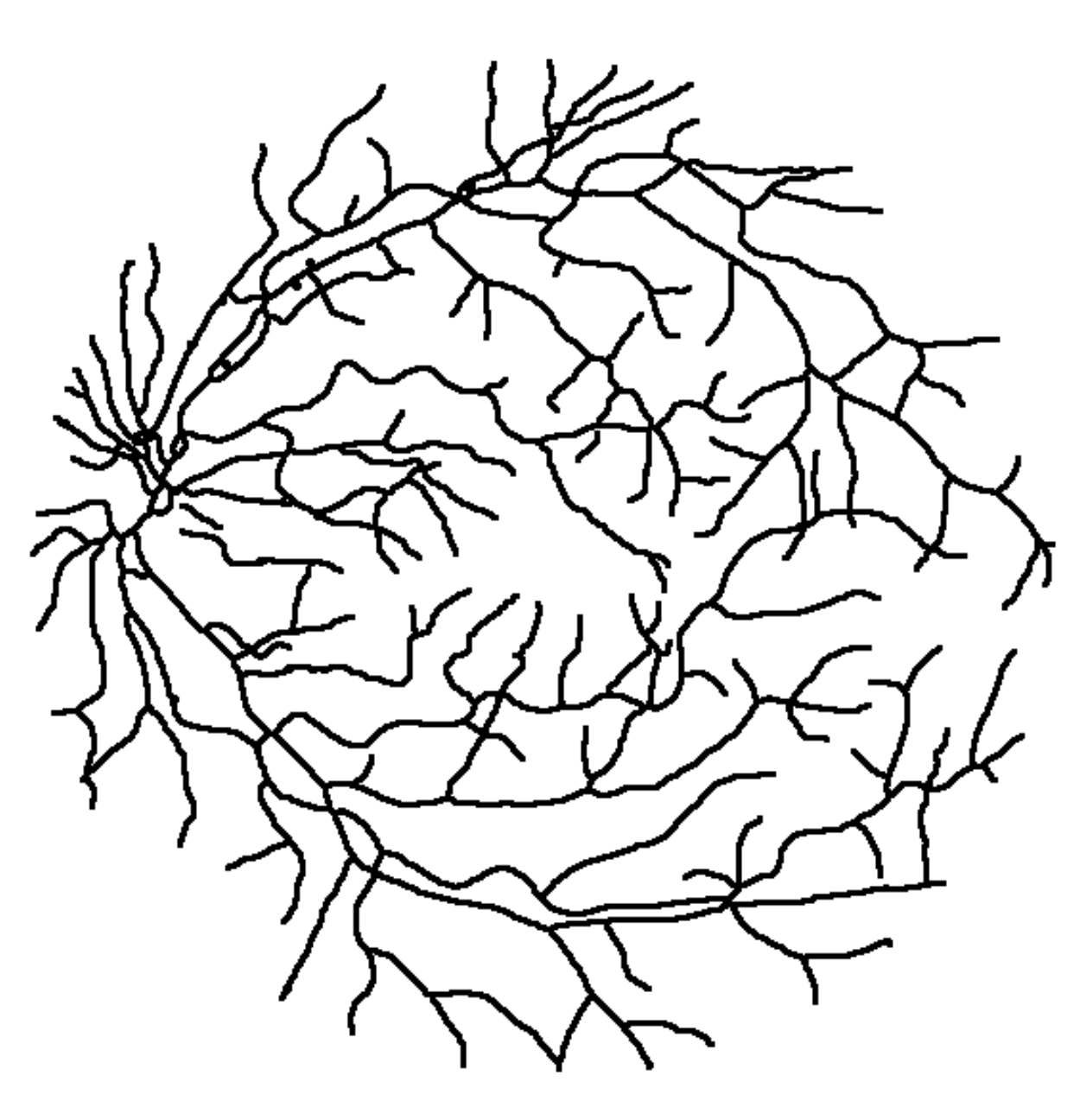}
		\caption{Skeleton}\label{fig:FullRetinaSkel}
	 \end{subfigure}
	 \hspace{-0.1in}
	\begin{subfigure}[b]{0.2\textwidth}
		 \includegraphics[width = 1.1\columnwidth]{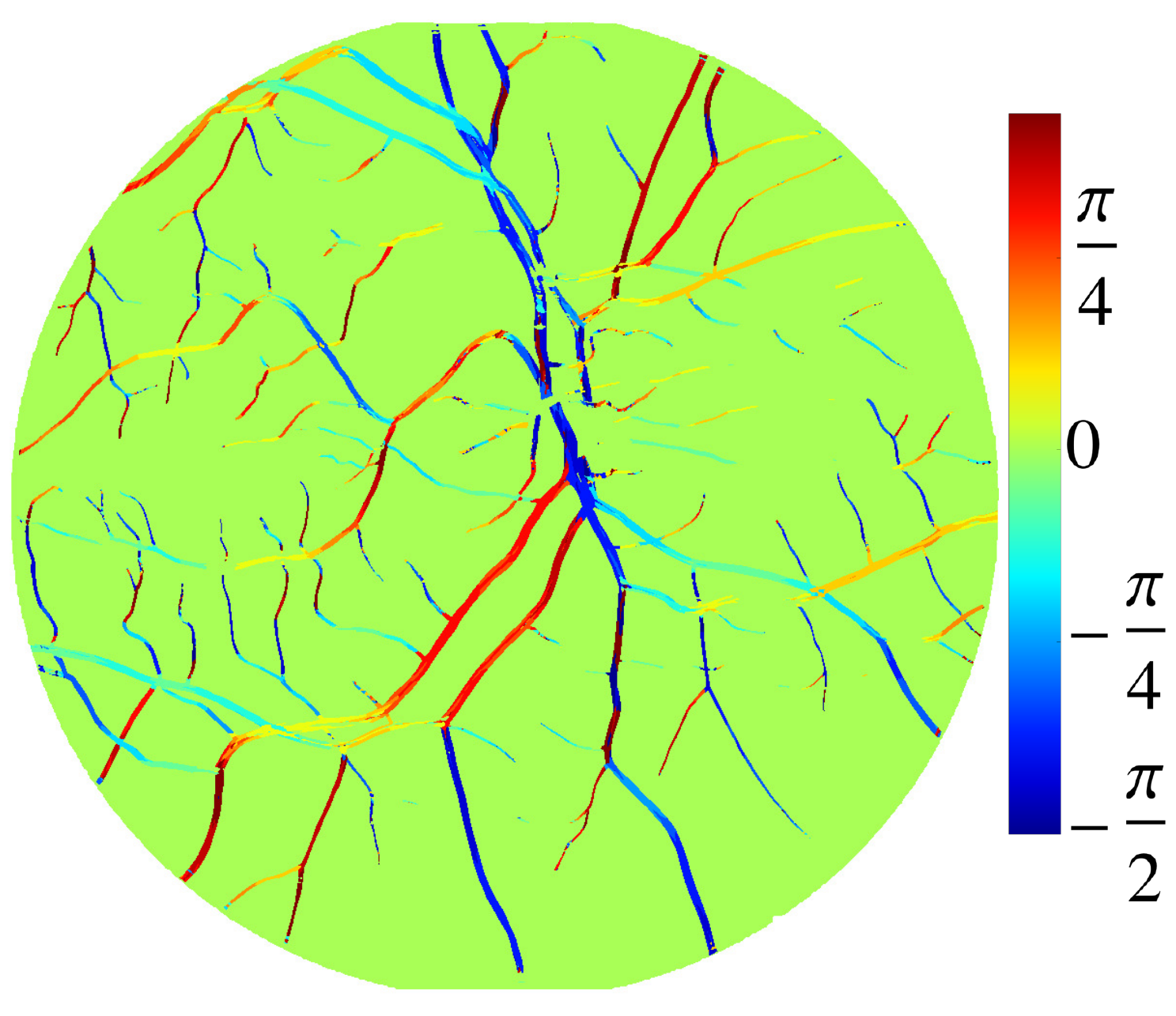} 
		 \includegraphics[width = 1.14\columnwidth]{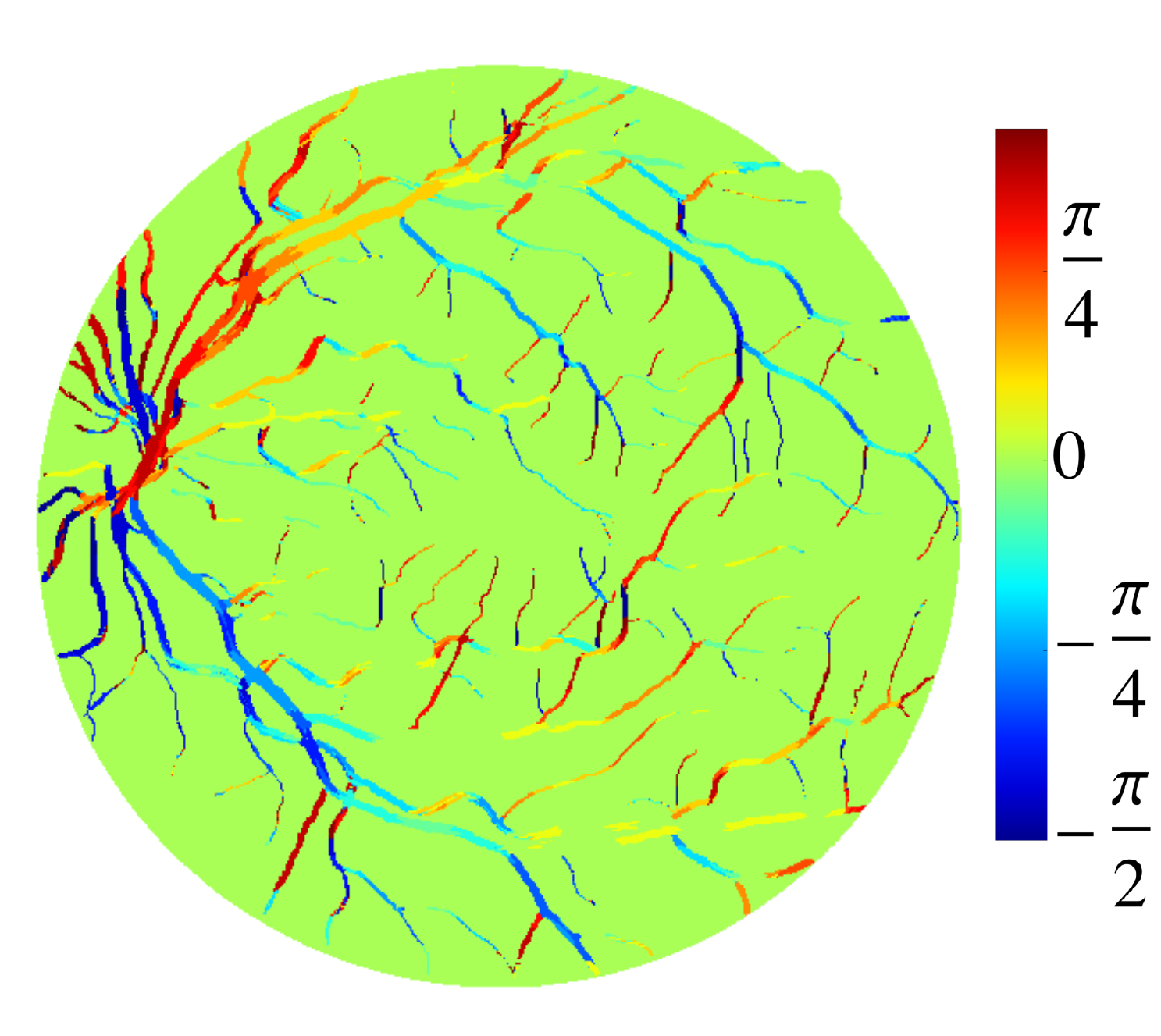}
		  \caption{Orientations} \label{fig:FullRetinaOSmap}
	\end{subfigure}

 \caption{Two sample images from the IOSTAR (top row) and the DRIVE (bottom row) datasets. (\subref{fig:FullRetinaOrig}) The original images, (\subref{fig:FullRetinaGT}) the vessel and (\subref{fig:FullRetinaAV}) the AV ground truth images (artery in red and vein in blue), (\subref{fig:FullRetinaSkel}) the vessel skeleton and (\subref{fig:FullRetinaOSmap}) the color-coded orientation maps}
  \label{fig:materials}
\end{figure*}

\sloppy Two retinal datasets have been used in this study.  
The public DRIVE dataset~\citep{staal2004} including 40 color fondus images  taken with a Canon CR5 non-mydriatic 3CCD camera, with a resolution of $565 \times 584$, a pixel size of $25\mu m/px$ and a field of view of $45^\circ$. The second dataset is the public IOSTAR dataset\footnote{Available at: \url{http://www.retinacheck.org/datasets}}~\citep{abbasi2015biologically} including 24 images taken with a scanning laser camera (SLO) with a resolution of $1024 \times 1024$, a pixel size of $14 \mu m/px$, and a field of view of $45^\circ$. The vessel ground truth and the AV labels are available for both datasets. 
Fig.~\ref{fig:materials} shows two sample images (\ref{fig:FullRetinaOrig}) from these two datasets together with their vessel (\ref{fig:FullRetinaGT}) and AV ground truth images (\ref{fig:FullRetinaAV}). The skeletons extracted from the vessel ground truth images are also presented there (\ref{fig:FullRetinaSkel}). The color-coded images in the last column (\ref{fig:FullRetinaOSmap}) show the dominant angle at each pixel location (See Step~\ref{eq:MIPOS} of Algorithm~\ref{alg:stepsHist}).
%-----------------------------------------------------------
\subsection{The statistical kernels}
The statistical kernel as explained in Sect.~\ref{sec:edgeStatistics} is calculated for both datasets. The number of discrete orientations used is  $n_\theta=16$ and $d$ is set to $65$. For each image both the full vasculature ground truth and the AV-separated ground truth images have been used and at the end, two different 3D histograms ($K^{stat}$) are obtained per dataset. The histograms extracted directly from the full vasculature network are called $K^{stat}_{DR}$ and $K^{stat}_{IO}$ and the ones obtained from the AV separated datasets are called $K^{stat}_{DR-AV}$ and $K^{stat}_{IO-AV}$. $DR$ stands for the DRIVE, $IO$ stands for the IOSTAR dataset and $AV$ stands for AV separated.
%One of the 4D histograms obtained from the IOSTAR dataset (in Step~\ref{st:accu} of Algorithm~\ref{alg:stepsHist}) has been visualized in Fig.~\ref{fig:4DDrAV}. The depicted histogram is $l_1$-normalized. The orientations ($\theta_m$ and $\theta'_m$) are changing over rows and columns in this figure and each $131\times131$ square represents the values of $H_{4D}(\Delta x, \Delta y, \theta_m, \theta'_m)$ at a fixed pair of $\{\theta_m,\theta'_m\}$.  Only 8 out of 16 orientations with the step size of $\pi/8$ are depicted here. As seen in this figure, the information is mainly concentrated at small orientation differences (when $\theta_m$ is close to $\theta'_m$). In addition, the orientation invariance property is a correct assumption, since the values in each row are similar to its previous row when a shifting and rotation are applied. 

\begin{figure}[htbp]
   \centering
   \includegraphics[width =0.95 \columnwidth,trim=0.03in 0 0 0,clip=true]{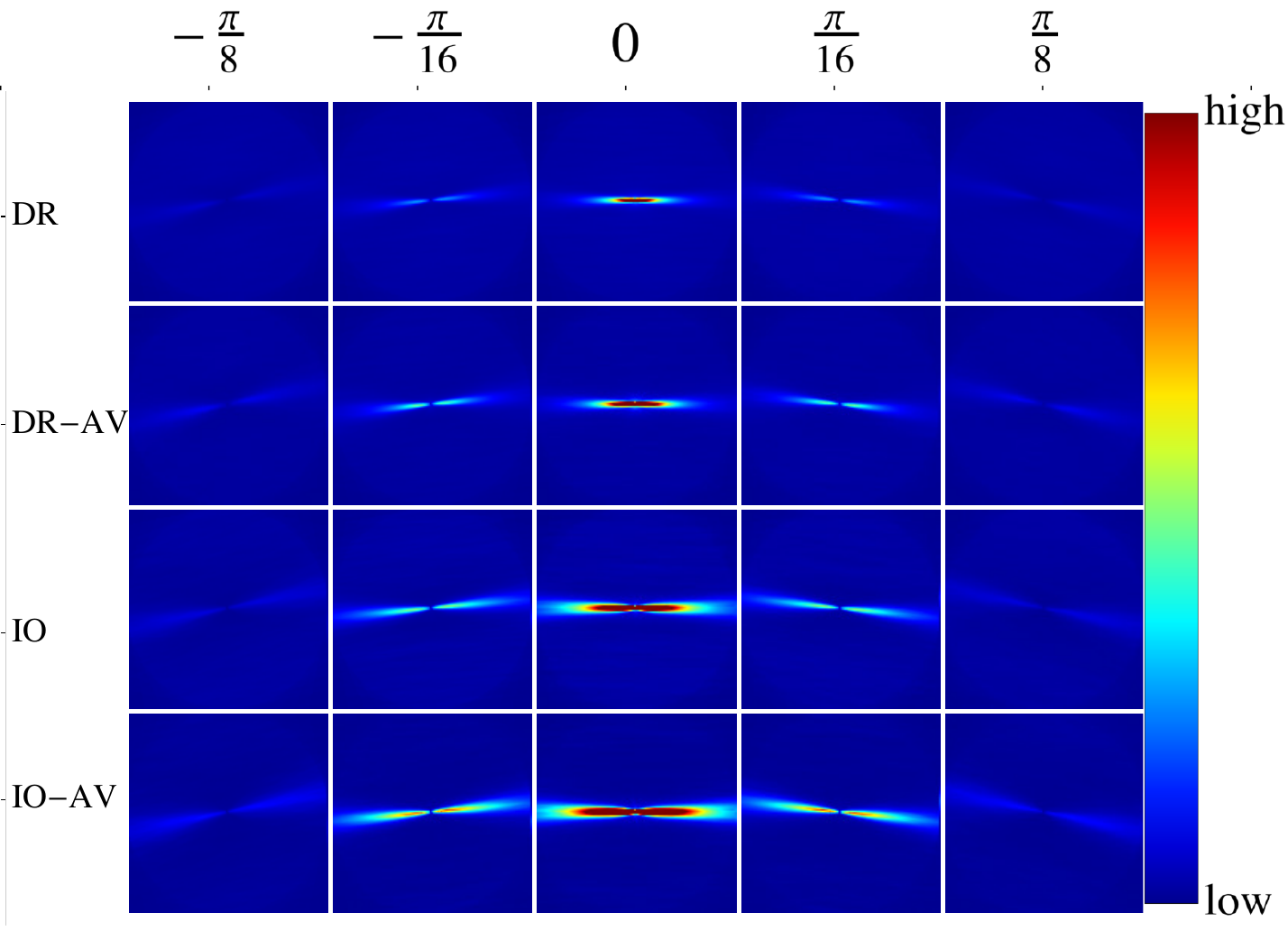} % requires the graphicx package
   \caption{The statistical kernels obtained in Step~\ref{st:final3Dhist} of Algorithm~\ref{alg:stepsHist} for each dataset. The value of $\theta$ is shown for each column and the values of all figures are clipped between 0 to 0.2 of the maximum value of the $K^{stat}_{DR}$}
   \label{fig:3Dhistograms}
\end{figure}

\begin{figure}[htbp]
  \centering 
 	 \begin{subfigure}[b]{0.45\columnwidth} \includegraphics[width =\columnwidth]{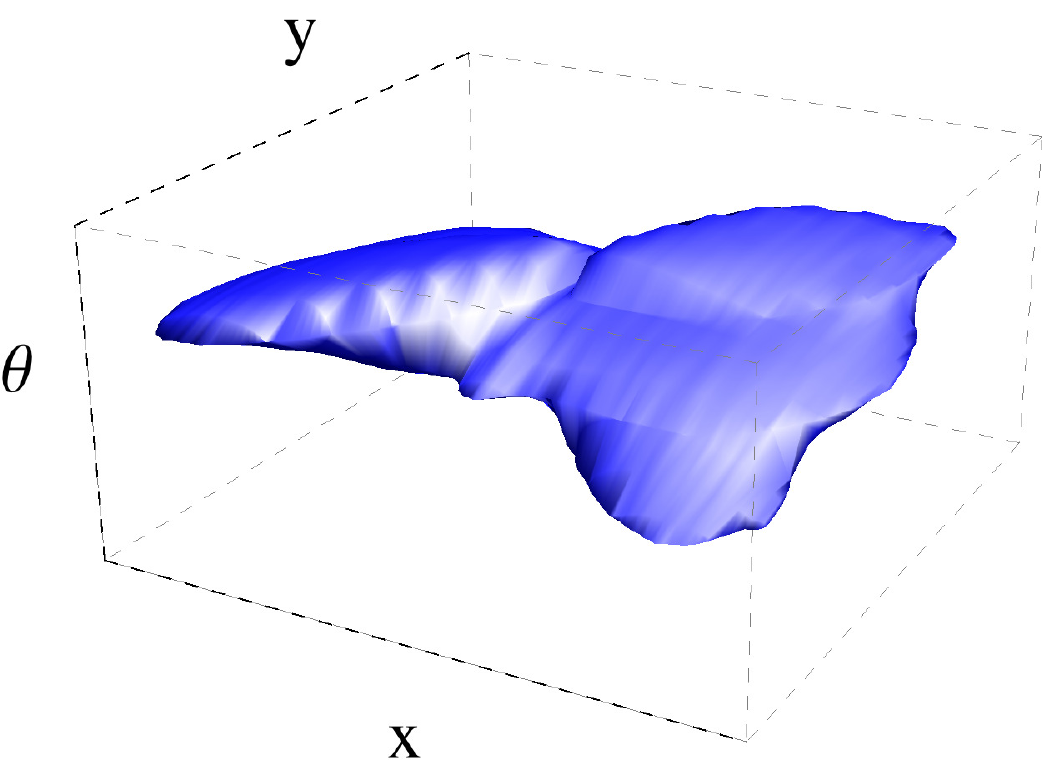} \caption{}\label{fig:isoDR}  \end{subfigure}
		\begin{subfigure}[b]{0.45\columnwidth}   \includegraphics[width = \columnwidth]{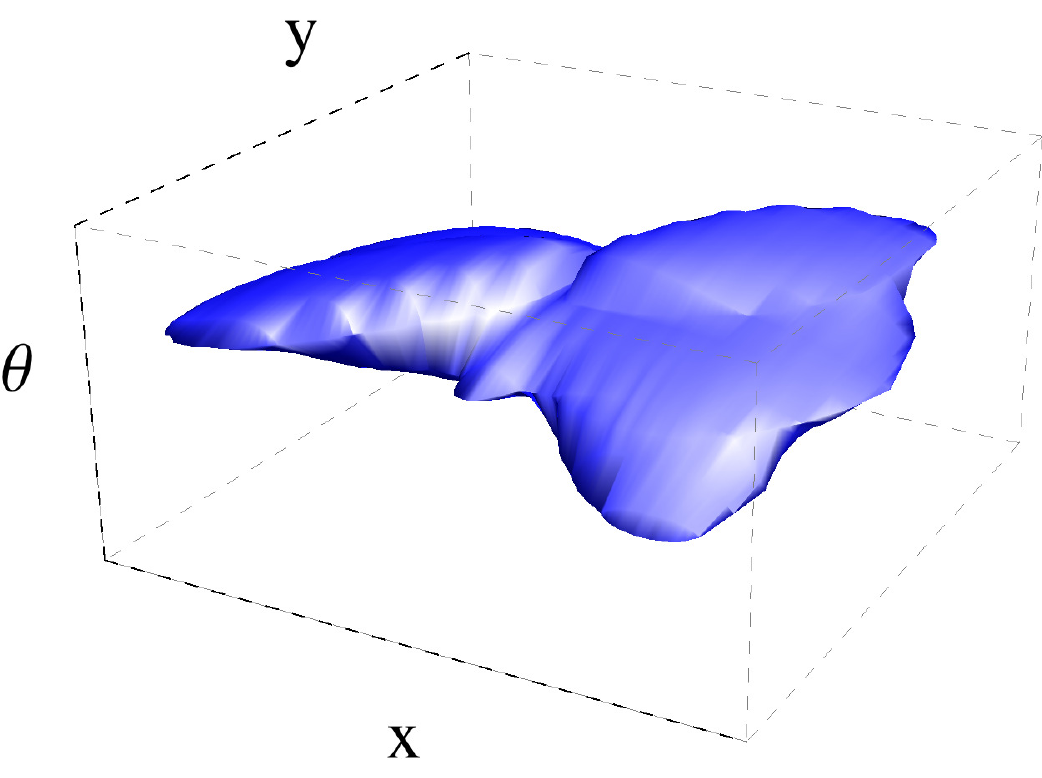} \caption{}  \label{fig:isoDRAV}\end{subfigure}
		 \begin{subfigure}[b]{0.45\columnwidth} \includegraphics[width = \columnwidth]{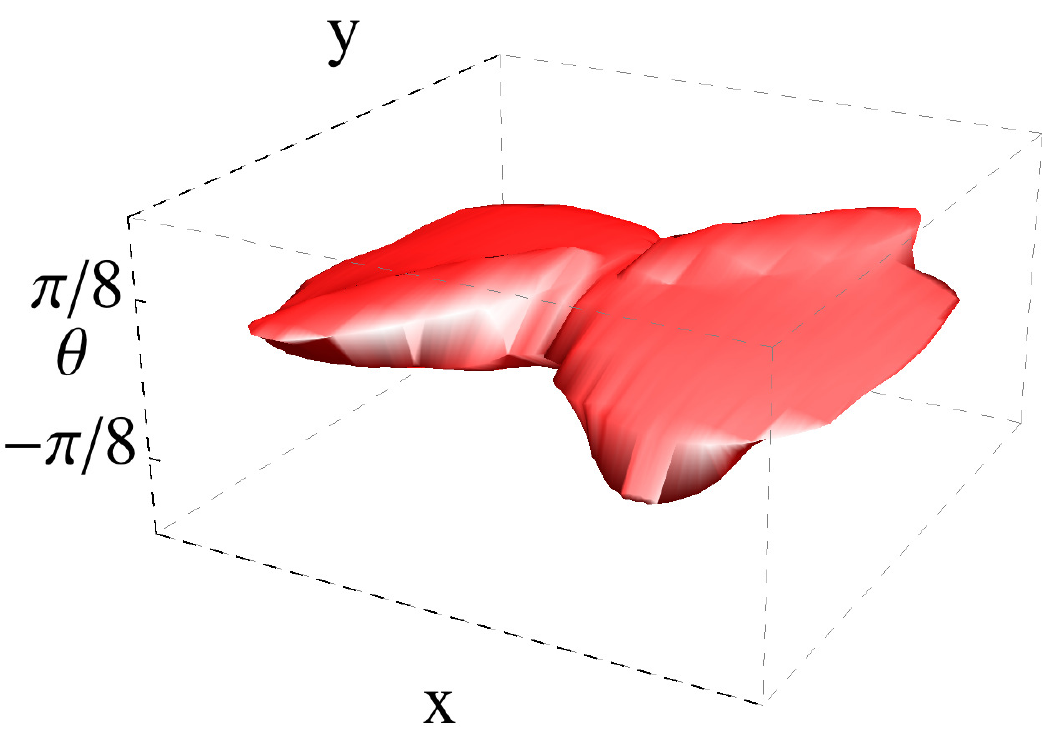}\caption{} \label{fig:isoIO} \end{subfigure}
		  \begin{subfigure}[b]{0.45\columnwidth} \includegraphics[width =\columnwidth]{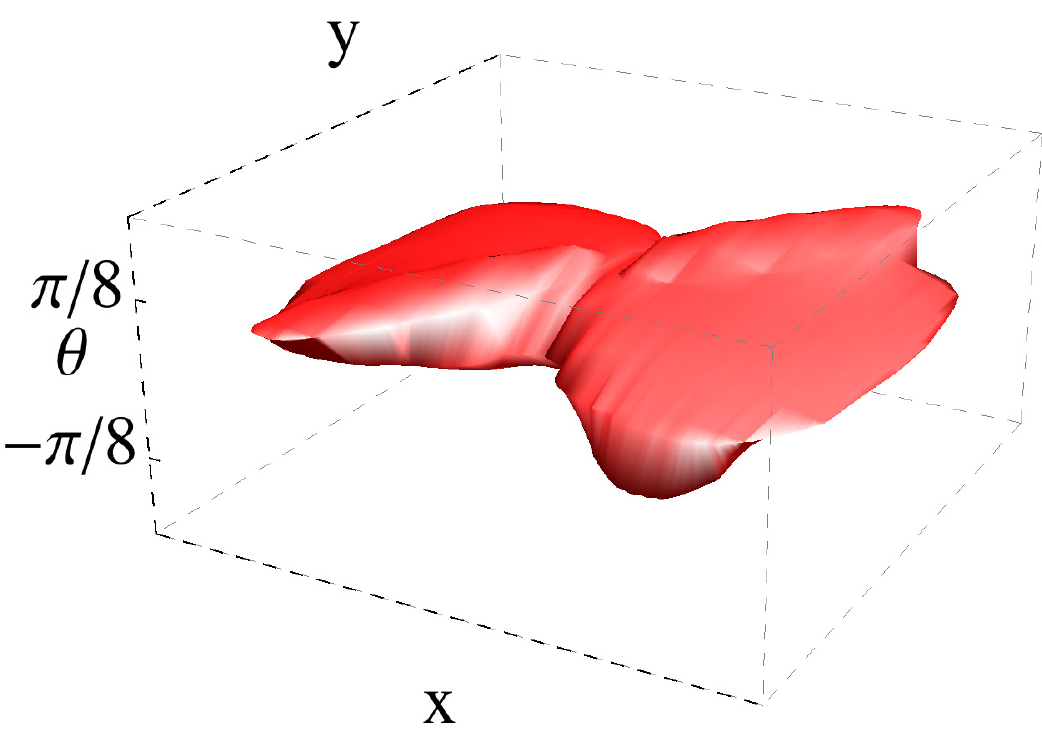} \caption{}\label{fig:isoIOAV} \end{subfigure}
  \caption{The level sets of the (\subref{fig:isoDR}) $K^{stat}_{DR}$ and (\subref{fig:isoDRAV}) $K^{stat}_{DR-AV}$ shown at 0.0125 of the maximum value of the $K^{stat}_{DR}$, and the levels sets of the (\subref{fig:isoIO}) $K^{stat}_{IO}$ and (\subref{fig:isoIOAV}) $K^{stat}_{IO-AV}$ shown at 0.025 of the maximum value of the $K^{stat}_{IO}$}
  \label{fig:isosurf}
\end{figure}

Two different visualizations of the final statistical kernels are shown in Fig.~\ref{fig:3Dhistograms} and~\ref{fig:isosurf}.  
%By making the histogram rotation-invariant (Step~\ref{st:3Dhist} of Algorithm~\ref{alg:stepsHist}), the 3D histograms are obtained. All the obtained histograms are $l_1$-normalized at the end. 
The rows in Fig.~\ref{fig:3Dhistograms} from top to bottom represent the $K^{stat}_{DR}$, $K^{stat}_{DR-AV}$, $K^{stat}_{IO}$, and $K^{stat}_{IO-AV}$ respectively. Each square has the dimension of $131\times 131$ and it depicts the value of the kernel at fixed relative orientation ($K^{stat}(\mathbf{x},\theta_c),~\theta_c\in\{\pm\pi/8,\pm\pi/16,0\}$). The kernel values of only five orientations are depicted as the information at other angles is very small. As seen in these figures, the maximum values of the statistical kernels occur at small orientation differences i.e., two aligned lines are more probable to appear in the images. This probability decreases when the orientation differences increase. Comparing these four histograms qualitatively, the statistical kernels of the DRIVE dataset are a bit less elongated compared to the ones of the IOSTAR datasets. Moreover, the separation of the lines of arteries and veins from each other results in less noisy histograms for both datasets; however, the difference is very small.
% In other words, two lines one belonging to an artery and the other one to the vein, should not be considered as a pair during the histogram calculation. 
 Fig.~\ref{fig:isosurf} visualizes the isosurfaces of these four data-driven kernels, which shows their high similarity.
%-----------------------------------------------------------
%\begin{figure}[htbp]
%   \centering
%   \includegraphics[width = 0.75\columnwidth,trim=0.1in 0 -0.2in 0,clip]{4DhistDrAV} % requires the graphicx package
%   \caption{The four dimensional histogram obtained from the IOSTAR-AV dataset in Step~\ref{st:accu} of Algorithm~\ref{alg:stepsHist}. The values at only 8 orientations are depicted. }
%   \label{fig:4DDrAV}
%\end{figure}
%-----------------------------------------------------------
%-----------------------------------------------------------
%\begin{figure}[htbp]
%   \centering
%   \includegraphics[width = 3in]{isosufIoAV} 
%   \caption{The level sets of the 3D histogram extracted from the IOSTAR (AV separated)}
%   \label{fig:3D}
%\end{figure}

These data-driven kernels are compared with each other quantitatively and their mutual differences are
 %least square errors ($l_2$-norm) between them is 
obtained as $error = \parallel K^{stat}_1 - K^{stat}_2 \parallel_{l_2} $, where $K^{stat}_1$ and $K^{stat}_2$ are the two statistical kernels in comparison. These least square errors are presented in Table~\ref{tbl:dataDrivenErrors} for each pair of kernels. Based on these quantitative results, their differences are very small. Therefore, it is possible to use them interchangeably, regardless of the dataset or the ground truth used for obtaining them. 
% Requires the booktabs if the memoir class is not being used
\begin{table}[htbp]
   \centering
   %\topcaption{Table captions are better up top} % requires the topcapt package
   \begin{tabular}{@{} lcr @{}} % Column formatting, @{} suppresses leading/trailing space
      \toprule
%      \multicolumn{2}{c}{Kernels} \\
%      \cmidrule(r){1-2} % Partial rule. (r) trims the line a little bit on the right; (l) & (lr) also possible
      $K^{stat}_1$    &  $K^{stat}_2$  & $error(\%)$\\
%      \midrule
	\hline
%      $K^{stat}_{DR}$      & $K^{stat}_{DR-AV}$ & 0.51\\
%     $K^{stat}_{IO}$        & $K^{stat}_{IO-AV}$     &  1.01 \\
%      $K^{stat}_{DR}$      & $K^{stat}_{IO}$  & 1.14 \\
%      $K^{stat}_{DR-AV}$      & $K^{stat}_{IO-AV}$  & 1.66 \\
	${DR}$      & ${IO}$  & 1.14 \\
      ${DR}$      & ${DR-AV}$ & 0.51\\
 	 ${DR}$      & ${IO-AV}$  & 2.14 \\
     	${IO}$        & ${IO-AV}$     &  1.01 \\
      ${IO}$        & ${DR-AV}$     &  0.69 \\
      ${DR-AV}$      & ${IO-AV}$  & 1.66 \\
      \bottomrule
   \end{tabular}
   \caption{The least square errors obtained by comparing each pair of the statistical kernels}
   \label{tbl:dataDrivenErrors}
\end{table}
%-----------------------------------------------------------
\subsection{Comparison to the probability model}
In this section we find the best approximations of the statistical kernels by comparing them against various probability kernels (obtained using Eq.~\ref{eq:2PitoPi})  with different parameters.
% which help in modelling the kernels numerically in case the data-driven kernels are not available. 
%by creating a large set of numerical kernels with 
The kernels that result in the least square errors are considered as the best approximations. 
%The Fourier-based technique (FBT) is used for creating the numerical solution and the kernels are compared in the spatial domain. 
%The varying parameters in FBT are $\alpha$ and $D_{33}$, which determine the expected life time of the resolvent kernel ($E(T)=1/\alpha$) and the diffusion matrix ($D = diag\{0,0,D_{33}\}$ ) respectively~\citep{zhang2014numerical}. It is worth mentioning that $D_{33} =\sigma^2/2$, in which $\sigma$ is the diffusion constant used in the Kolmogorov equation of Mumford's direction process~\citep{zhang2014numerical,mumford1994elastica}.
%
In our experiments, the parameter $\alpha$ takes 50 different values from 0.00001 to 0.01 and the parameter $D_{33}$ takes 100 values from 0.000001 to 0.005. These ranges are determined heuristically. The minimum errors and the corresponding parameters for each kernel are presented in Table~\ref{tbl:CompErrors}. Based on these results, the errors are very small for all the kernels. The largest error is related to the $K^{stat}_{IO-AV}$, which is close to $1\%$. 
%A related point to consider is that the diffusion constants obtained for the best approximations of the statistical kernels are very close to the diffusion constant of the stochastic kernel found heuristically by~\cite{favali2016analysis}.
% Requires the booktabs if the memoir class is not being used
\begin{table}[htbp]
   \centering
   \caption{The least square errors and the corresponding parameters resulting in these errors between the statistical and probabaility kernels}
  \begin{tabular}{@{\extracolsep{1pt}} c*{4}{c} } 
  	 \toprule
	$K^{stat}$	&	error ($\%$) & $\alpha$ & $D_{33}$ & $\sigma$\\
	 \hline
%	\midrule
%	 $K^{stat}_{DR}$ 	&	0.3275	&	0.0024	&	0.00170	&	0.0583 \\
%	 $K^{stat}_{DR-AV}$ &	0.55 		&	0.0048	&	0.00210	&	0.0648\\
%	$K^{stat}_{IO}$	&	0.8109	&	0.0080	&	0.00130	&	0.0509 \\
%	$K^{stat}_{IO-AV}$ & 	1.04 		& 	0.0098 	& 	0.00085	& 	0.0412\\
	${DR}$ 	&	0.3275	&	0.0024	&	0.00170	&	0.0583 \\
	 ${DR-AV}$ &	0.55 		&	0.0048	&	0.00210	&	0.0648\\
	${IO}$	&	0.8109	&	0.0080	&	0.00130	&	0.0509 \\
	${IO-AV}$ & 	1.04 		& 	0.0098 	& 	0.00085	& 	0.0412\\
     \bottomrule
    \end{tabular}
   \label{tbl:CompErrors}
\end{table}

\begin{figure}[htbp]
  \centering 
 	 	\begin{subfigure}[b]{0.45\columnwidth} \includegraphics[width = \columnwidth]{isosufDr}  \caption{}\label{iso-DR} \end{subfigure}
		\begin{subfigure}[b]{0.45\columnwidth} \includegraphics[width = \columnwidth]{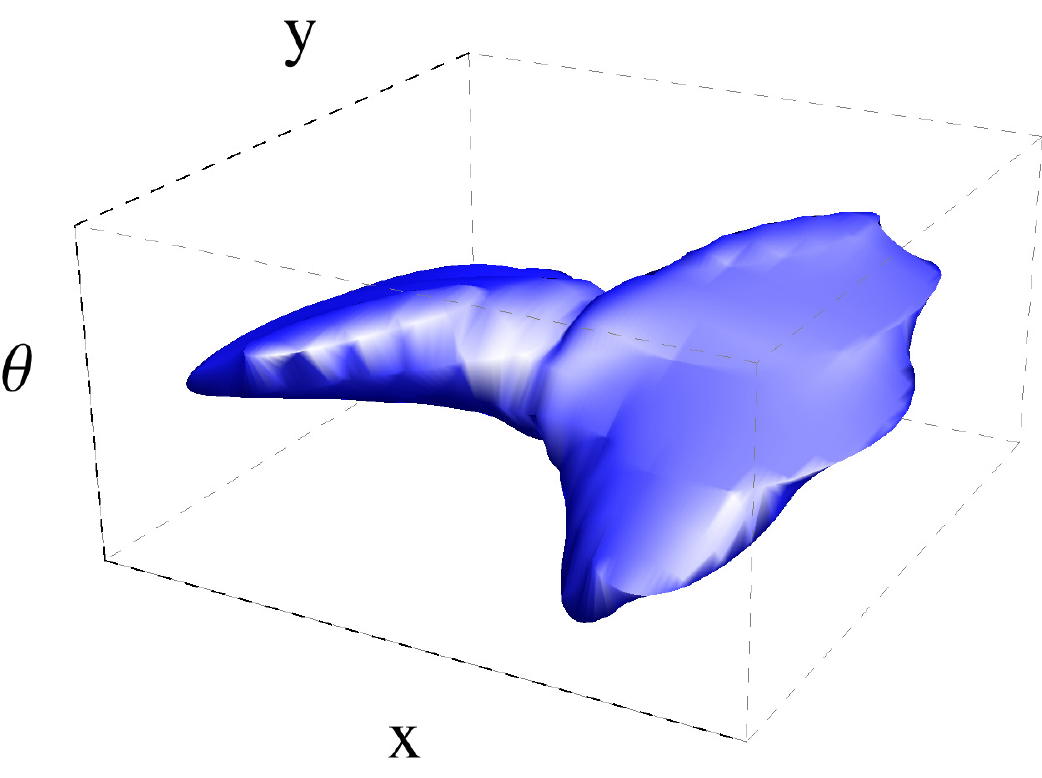}  \caption{}\label{iso-DR-FBT} \end{subfigure}\\
		 \begin{subfigure}[b]{0.55\columnwidth} \includegraphics[width =\columnwidth]{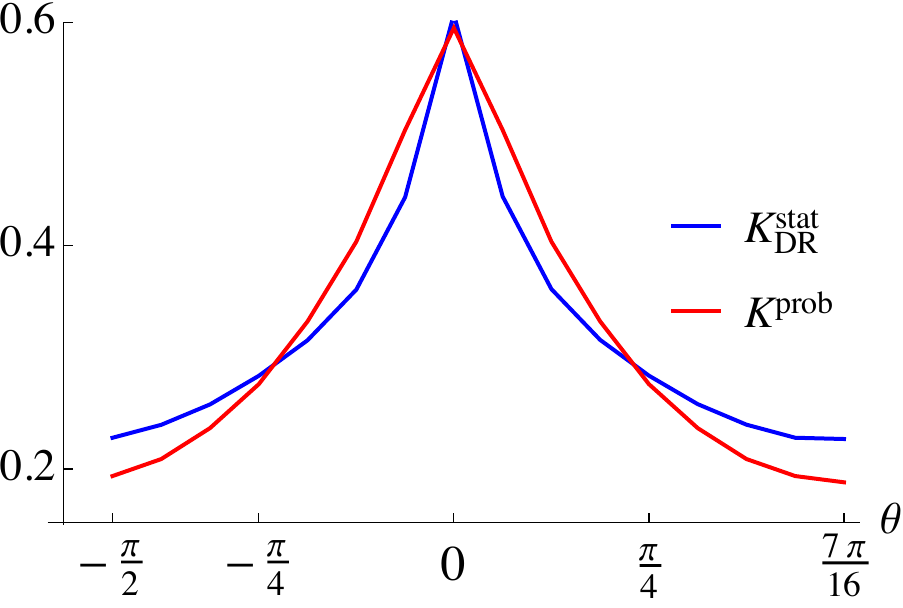} \caption{}\label{sumxyFBT-DR} \end{subfigure}
\begin{subfigure}[b]{1\columnwidth}   \includegraphics[width = \columnwidth]{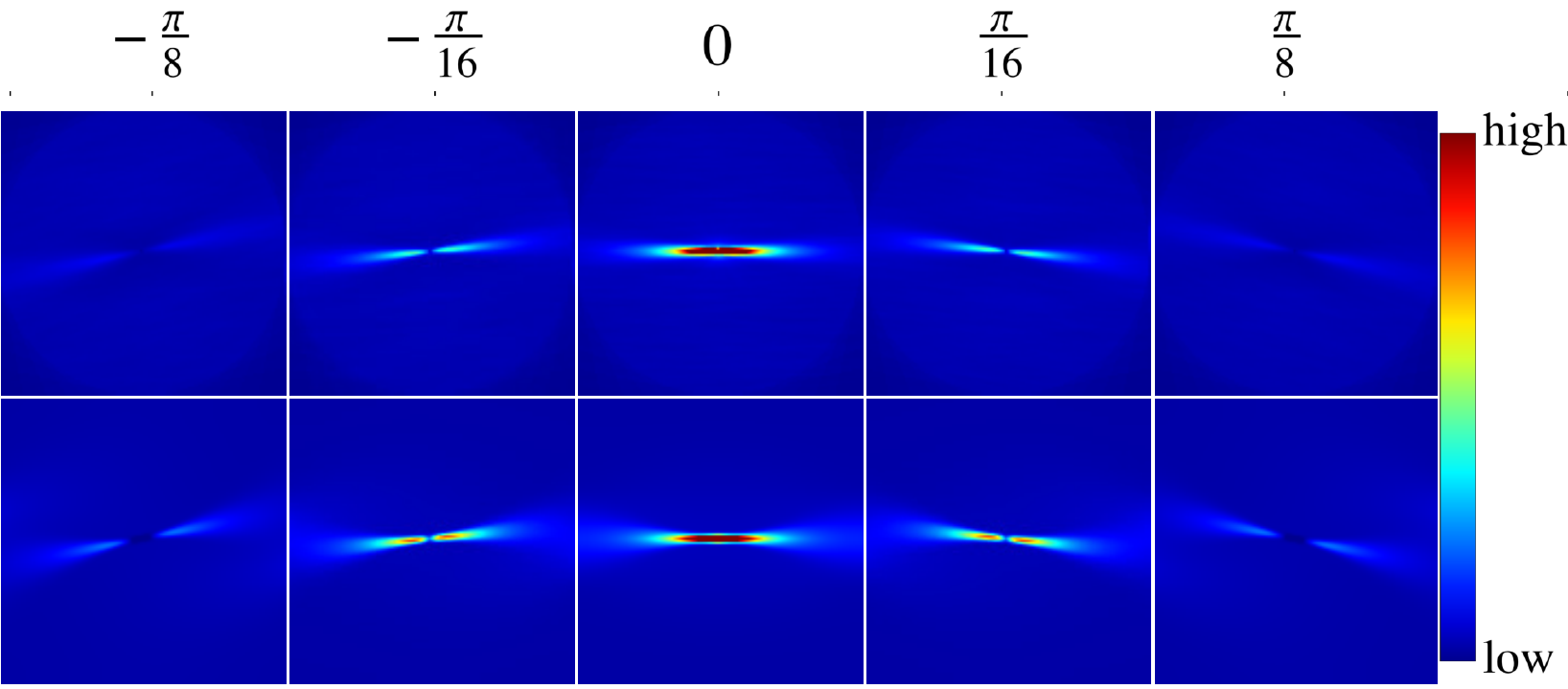} \caption{}  \label{DR-FBT}\end{subfigure}
%		 \begin{subfigure}[b]{1.6in} \includegraphics[width = 1.7in]{} \label{}\caption{IOSTAR} \end{subfigure}
		 
  \caption{Comparison between the $K^{stat}_{DR}$ and its best approximating kernel $K^{prob}_{DR}$: the level sets at 0.012 of the maximum value of (\subref{iso-DR}) the statistical and (\subref{iso-DR-FBT}) the probability kernels, (\subref{sumxyFBT-DR}) the summation of the values of the two kernels over the spatial dimension at different angles, and (\subref{DR-FBT}) the line distribution at five different orientations $\theta\in\{\pm\pi/8,\pm\pi/16,0\}$ in the data-driven (top row) and the probability kernel (bottom row)}
  \label{fig:DR-FBT-visual}
\end{figure}

%Since the four statistical kernels are very similar to each other, we only show 
A sample qualitative comparison of these two types of kernels is shown in Fig.~\ref{fig:DR-FBT-visual}. In addition to very similar profiles of the two kernels in Fig.~\ref{iso-DR}, \subref{iso-DR-FBT} and \subref{DR-FBT}, the summations of the line distribution over the spatial dimension also matches the information density of the probability kernel at every $\theta$ layer (Fig.~\ref{sumxyFBT-DR}). In both kernels, the maximum value appears at $\theta=0$ as expected.
%-----------------------------------------------------------
\subsection{Application in retinal image analysis}
In this section, the application of the statistical model of the cortical connectivity pattern in identifying vessel connections in retinal images is explained. 
%Aforementioned, in contrast to the method proposed by \cite{favali2016analysis}, the data-driven kernel is used as the cortical connectivity kernel. 

In this method the vessel connections are retrieved from image segmentations (not necessarily centerlines). So an initial segmentation of the image $I$ (using the aforementioned segmentation techniques proposed in the literature) is required. The binary image representing the segmentation is called $I_s$. Repeating the steps as the ones proposed in Algorithm~\ref{alg:stepsHist} (Steps~\ref{st:OST1},~\ref{eq:MIPOS} and \ref{st:intSet}) the image is lifted ($U_I$), dominant orientations ($\theta_m$) are obtained and the set of interest points is created as: $$L= \{(\mathbf{x},\theta_{m}(\mathbf{x}))~|~I_{s}(\mathbf{x}) = 1\}.$$ Considering $m = |L|$, an $m\times m$ affinity matrix ($A$) is created to take into account the connection probability between each pair of points in set $L$ and is later aggregated with another affinity matrix ($\tilde{A}$) representing the similarities of the corresponding vessel intensities. So for each pair of $(\mathbf{x}_i,\theta_m(\mathbf{x}_i))$ and $(\mathbf{x}_j,\theta_m(\mathbf{x}_j))\in L$:
%\begin{equation}
%\begin{split}
%A(\mathbf{x}_i,\mathbf{x}_j) = k^{stat}((\mathbf{x}_i,\theta_m(\mathbf{x}_i)),(\mathbf{x}_j,\theta_m(\mathbf{x}_j)),\\ i,j=1,\dots,m.
%\end{split}
%\end{equation}
%This affinity matrix is aggregated with another similarity matrix ($\tilde{A}$) representing the similarities between the vessel intensities as:
\begin{equation}
\begin{split}
A_{final}(\mathbf{x}_i,\mathbf{x}_j) = A(\mathbf{x}_i,\mathbf{x}_j) \tilde{A}(\mathbf{x}_i,\mathbf{x}_j)=\\
 k^{stat}((\mathbf{x}_i,\theta_m(\mathbf{x}_i)),(\mathbf{x}_j,\theta_m(\mathbf{x}_j)) \times \\ 
 G_{\sigma_{int}}(I_n(\mathbf{x}_i)-I_n(\mathbf{x}_j))
%  \exp\frac{(I_n(\mathbf{x}_i)-I_n(\mathbf{x}_j)}{\sigma_{int}}
  ,\\ i,j=1,\dots,m.
\end{split}
\end{equation}
where $G_{\sigma_{int}}(x)$ is the normalized Gaussian kernel with the standard deviation of $\sigma_{int}$ and $I_n$ is the image intensity in normalized green channel. Here we normalized luminosity and contrast using the method by~\cite{Foracchia2005}.
%The Gaussian kernel has been applied for creating this similarity matrix (see~\cite{favali2016analysis} for more details).  
The final affinity matrix is analyzed using the self-tuning spectral clustering technique, which identifies the salient groups in the image automatically. 
%A sample of expected clusters, detected by our visual perception easily, is shown in three colors in Fig.~\ref{fig:phGT}. 
%\begin{figure}[htbp]
%  \centering 
%  		\begin{subfigure}{0.2\columnwidth} \includegraphics[width =\columnwidth]{bifCross_seg}\caption{} \label{fig:phSeg}  \end{subfigure}
%		\qquad
%		\begin{subfigure}{0.2\columnwidth} \includegraphics[width =\columnwidth]{bifCross_gt}  \caption{}\label{fig:phGT} \end{subfigure}
%	 \caption{(\subref{fig:phSeg}) A sample image with interrupted segments and (\subref{fig:phGT}) the contours identified in the image by our perceptual system.}
%  \label{fig:phantoms}
%\end{figure}

As discussed before, the differences between the kernels obtained from the same datasets are very small and they can be used interchangeably.  Hence, for analyzing the retinal image patches taken from the DRIVE and IOSTAR datasets we only use the $K^{stat}_{DR-AV}$ and the $K^{stat}_{IO-AV}$ kernels respectively. For testing the method, several patches with the sizes of $51\times 51$ and $101\times101$ have been selected around junctions from the DRIVE and IOSTAR datasets respectively. 
%{\color{magenta}Then using the method described in Sect.~\ref{},}
%the applicability of the data-driven kernels in detecting the vessel contours correctly have been investigated. 
 %
%
%The probability of having a connection is investigated only among the pixels which are determined to be vessel pixels in vessel segmentations. Therefore, 
%We aim to find groups formed by the pixels labeled as vessels. Thus, 
A vessel segmentation by~\cite{abbasi2015biologically} developed for both color RGB images and SLO images is applied on these patches. %This method is designed not only for the color RGB images, but also for removing the noise in the SLO images and segmenting them afterwards. This method was validated before on both datasets. 
After segmenting the images, the positions, orientations and normalized intensities for the vessel pixels are extracted. 
%Then for each pair of vessel pixels, an affinity matrix is created using the data-driven kernel. This affinity matrix represent the probability that the two vessel pixels with the specific positions and orientations belong to the same contour. This affinity matrix is later aggregated with another similarity matrix representing the similarities between the intensities of the vessel pixels. Finally, by applying the slef-tuning spectral clustering technique the salient groups are identified in the image (the reader is referred to~\cite{favali2016analysis}for more details about the method). 

\begin{figure*}[htbp]
  \centering 
        	 \vspace{-0.5in}
   	\begin{subfigure}{0.05\textwidth}  
		% trim=l b r t	
		\includegraphics[width =1.5\columnwidth,trim = 0 0.1in 0 -0.4in,clip]{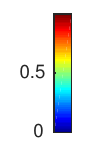}
		\vspace{0.1in}
		\includegraphics[width= 1.5\columnwidth,trim = 0 -0.1in 0 0in,clip]{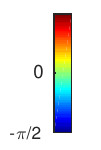}
		\vspace{1.65in}
   	 \end{subfigure}
 	 \begin{subfigure}{\sizp}  
	 	\includegraphics[width = \sizz]{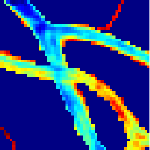} 
		\includegraphics[width = \sizz]{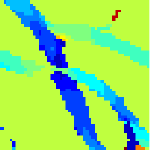}
		\includegraphics[width = \sizz]{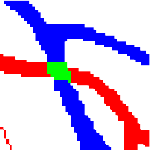}
		\includegraphics[width = \sizz]{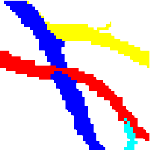}
	\caption{$D_1$} \label{fig:D1}
	 \end{subfigure}
	  \begin{subfigure}{\sizp} 
	 	\includegraphics[width = \sizz]{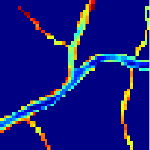}  
		\includegraphics[width = \sizz]{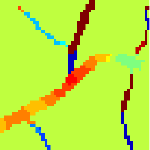}
		\includegraphics[width = \sizz]{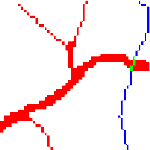}
		\includegraphics[width = \sizz]{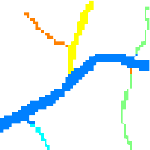} 
	\caption{$D_2$} \label{fig:D2}
	 \end{subfigure}
	  \begin{subfigure}{\sizp} 
	 	\includegraphics[width = \sizz]{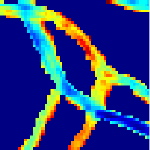} 
		\includegraphics[width = \sizz]{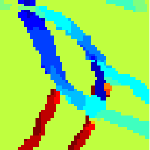} 
		\includegraphics[width = \sizz]{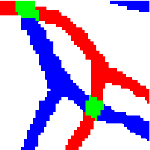}
		\includegraphics[width = \sizz]{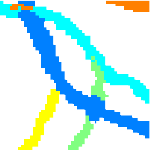}
	\caption{$D_3$} \label{fig:D3}
	 \end{subfigure}
	  \begin{subfigure}{\sizp} 
	 	\includegraphics[width = \sizz]{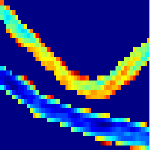} 
		\includegraphics[width = \sizz]{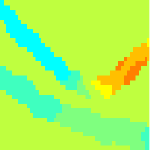}
		\includegraphics[width = \sizz]{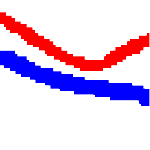}
		\includegraphics[width = \sizz]{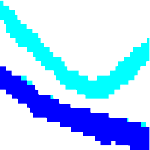} 
	\caption{$D_4$} \label{fig:D4}
	 \end{subfigure}
	 \begin{subfigure}{\sizp} 
	 	\includegraphics[width = \sizz]{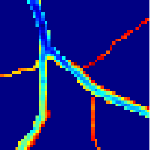} 
		\includegraphics[width = \sizz]{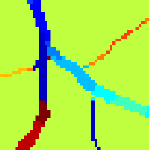}
		\includegraphics[width = \sizz]{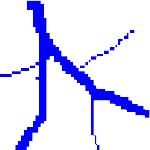}
		\includegraphics[width = \sizz]{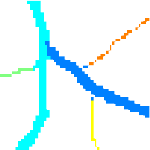} 
	\caption{$D_5$} \label{fig:D5}
	 \end{subfigure}
	 \begin{subfigure}{\sizp} 
	 	\includegraphics[width = \sizz]{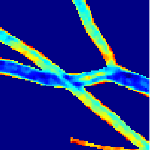} 
		\includegraphics[width = \sizz]{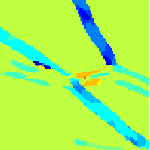}
		\includegraphics[width = \sizz]{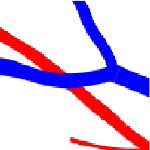}
		\includegraphics[width = \sizz]{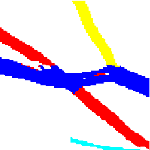} 
	\caption{$I_1$} \label{fig:I1}
	 \end{subfigure}
	 \begin{subfigure}{\sizp} 
	 	\includegraphics[width = \sizz]{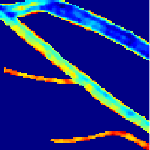} 
		\includegraphics[width = \sizz]{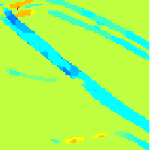}
		\includegraphics[width = \sizz]{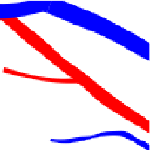}
		\includegraphics[width = \sizz]{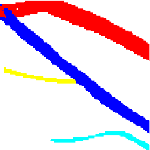} 
	\caption{$I_2$} \label{fig:I2}
	 \end{subfigure}
	 \begin{subfigure}{\sizp} 
	 	\includegraphics[width = \sizz]{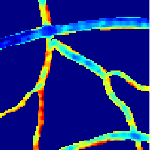} 
		\includegraphics[width = \sizz]{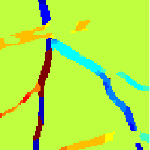}
		\includegraphics[width = \sizz]{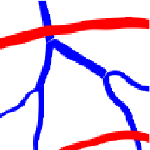}
		\includegraphics[width = \sizz]{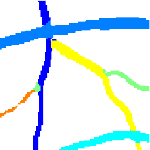}
	\caption{$I_3$} \label{fig:I3}
	 \end{subfigure}
	 \begin{subfigure}{\sizp} 
	 	\includegraphics[width = \sizz]{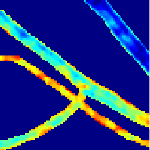} 
		\includegraphics[width = \sizz]{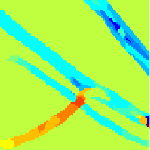}
		\includegraphics[width = \sizz]{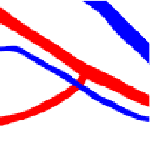}
		\includegraphics[width = \sizz]{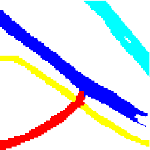} 
	\caption{$I_4$} \label{fig:I4}
	 \end{subfigure}
	 \begin{subfigure}{\sizp} 
	 	\includegraphics[width = \sizz]{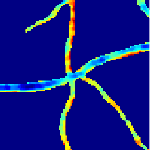} 
		\includegraphics[width = \sizz]{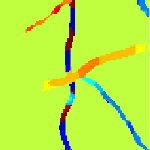}
		\includegraphics[width = \sizz]{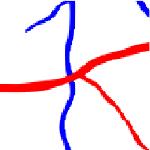}
		\includegraphics[width = \sizz]{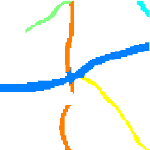} 
	\caption{$I_5$} \label{fig:I5}
	 \end{subfigure}
	 \vspace{-0.5in}
		 %%
%	\begin{subfigure}{\sizz}   \includegraphics[width = \sizp]{D1-osmap}\caption{}   \label{fig:os}\end{subfigure}
%	%%
%	 \begin{subfigure}{\sizz} \includegraphics[width = \sizp]{D1-avGT}\caption{}  \label{fig:av}\end{subfigure}
%	 %%
%	\begin{subfigure}{\sizz} \includegraphics[width = \sizp]{D1-res} \caption{} \label{fig:res}\end{subfigure}
%	%%
  \caption{Some example retinal patches selected from the DRIVE ($D_1$ to $D_5$) and the IOSTAR ($I_1$ to $I_5$). The first three rows from top to bottom indicate the intensity, main orientations and the AV labels of the vessels. The last row represents the final clustering results after removing the small groups (noise). Detected clusters are shown in different colors}
  \label{fig:clustRes}
\end{figure*}

Fig.~\ref{fig:clustRes} shows five image patches for each dataset. The first five patches ($D_1$ to $D_5$) belong to the DRIVE and the second five patches ($I_1$ to $I_5$) are selected from the IOSTAR datasets. For each patch, from top to bottom the vessel intensities, the color-coded orientation maps (representing the values of $\theta_m(\mathbf{x})$), the AV labels,  and finally the clustering results have been presented. As seen in this figure, the detected final perceptual units (shown in different colors) correspond to the separate blood vessels existing in the image patch. These patches have been selected in a way to represent several complex topological structures of the blood vessels, varying the number of the vessel bifurcations and crossings and existence of parallel or curved vessels. For all the experiments the parameter $\sigma_{int}$ was set to $0.2$. Despite the complexity of the structures, using orientation and intensity features for determining the contextual connections among pixels, individuates well the blood vessels from each other. Using the feature of intensity helps in the cases where there is an abrupt change in the orientation but not the intensity e.g. in $D_3$ and $D_4$. Moreover, the presence of disconnections (e.g. in $I_5$) does not affect the correct grouping. In addition to these patches, 20 more patches per dataset have been analyzed. For all these cases, the method is successful in correctly grouping the vessel pixels. The limitation arises when both the feature of intensity and orientation of a vessel are very noisy or change suddenly. In these cases, the vessel splits into smaller clusters. Involving additional contextual information such as curvature or scale may help in resolving this problem as proposed by~\cite{abbasi2016cortically}. 

%In general, the data-driven kernels are sufficient for analyzing these patches and it is not needed to create the kernel for each patch separately. It helps in skipping the exhaustive parameter tuning for creating the numerical kernels and avoids additional optimization steps for finding the best parameters. Moreover, using the self-tuning spectral clustering is another advantageous step toward automatic selection of groups and the eigenvalue thresholding step is not needed anymore.
%-----------------------------------------------------------
%\subsection{Discussion}
%{\color{magenta}\begin{itemize}
%\item discussion about comparison results and the differences between the two datasets and their parameters
%\item Discussing the clustering results and discussion about the application of data-driven statistics in perceptual completion
%\end{itemize}}
%
%%-----------------------------------------------------------

% !TEX root = EdgePaper.tex
\section{Conclusion}
\label{sec:conclusion}
In this work, we exploit the relation between the statistical co-occurrences of line elements in natural images and the high level task of contour grouping in our brain, and use it in the retinal image processing for contour completion. Firstly, the steps for obtaining the rotation and translation-invariant line statistics from different retinal image datasets are explained. Secondly, their relation to the symmetrized probability kernel of the direction process on the projective line bundle modeling the cortical connectivity is investigated. The results reveal their remarkable high similarity. However, in edge co-occurrences (rather than line co-occurrences) of natural images \citep[e.g.][]{sanguinetti2010model} the shapes seem to resemble the shape of the hypo-elliptic heat kernels (without convection) in $\mathbb{R}^2\times P^1$, but the relation needs further investigation.
%They show that the contour grouping capability of the brain is directly related to the statistical co-occurrence of edge elements in natural images. 
%
%The fundamental solution of Fokker-Planck equation has been widely used to represent the geometrical structure of multi-orientation cortical connectivity in V1. Several numerical implementations for creating the fundamental solution of FP equation has been proposed in the literature addition to the exact solution. Since the Fourier-based technique provides the best approximate of the exact solution, it was employed in this work to create the connectivity kernel for a large set of parameters. Then by comparison between the data-driven kernels and the numerical kernels, the best matches have been selected. In this way the parameters modelling the edge co-occurrence statistics are obtained. These parameters can be used in future parameter tuning in other computer vision algorithms using these kernels.
In addition, all the statistical kernels obtained from different datasets are compared with each other quantitatively and qualitatively. The obtained results indicate their high similarity and reproducibility despite the differences in the datasets and the setups used. 
%Separation of the contours of arteries from veins before obtaining the cross-correlations resulted in less noisy and more accurate histograms.  

Furthermore, the statistical kernels were used directly to retrieve the individual vessels from segmented retinal images. The successful results show that Mumford's direction process is a very good model for centerlines of vessels, and together with the Lie group theory, the proposed connectivity analysis technique is useful for retinal image analysis.  
%instead of the stochastic connectivity kernel in a previously introduced method for analyzing the vessel connections in retinal images~\citep{favali2016analysis}. 
Using the data-driven kernel (that does not need parameter tuning) in addition to adding the automatic self-tuning spectral clustering technique, forms a robust and fully automatic connectivity analysis technique. This provides an effective solution for challenging situations in which most of the methods fail (but our visual perception succeeds) because of non-perfect imaging conditions, interruptions, or occlusions.    
%The only remaining parameter is the $\sigma_{int}$ (set to 0.2 in this work) used for adjusting the effect of feature of intensity in creating the affinity matrix. However, based on previous validation of the method and the experiments of this article, this parameter is almost constant and its variation is very small. %
%In general, the data-driven kernels are sufficient for analyzing these patches and it is not needed to create the kernel for each patch separately. It helps in skipping the exhaustive parameter tuning for creating the numerical kernels and avoids additional optimization steps for finding the best parameters. Moreover, using the self-tuning spectral clustering is another advantageous step toward automatic selection of groups and the eigenvalue thresholding step is not needed anymore.

The method presented here for extracting line co-occurrences from retinal images and the improved connectivity analysis approach can be extended to a rich number of other application areas which contain curvilinear structures such as  corneal never fibers, plant roots and road networks. For each dataset, it is possible to learn the connectivity kernel and use it directly for similar images. 
%We used the vessel ground truth images and the centerline extracted from them to obtain the edge statistics in these images. However, in case the ground truth images are not available, a segmentation or centerline extraction technique can be used instead. 
%As mentioned in Sect.~\ref{sec:intro} several segmentation approaches have been proposed in the literature for the images containing these structures.

Since the visual cortex deploys additional contextual information and its receptive fields are not only sensitive to orientation, but also other information (such as scale and curvature), one potential extension of the method is to use this additional information in deriving the line statistics and creating higher order kernels. 

% !TEX root = EdgePaper.tex
%\section*{Acknowledgement}
\begin{acknowledgements}
This project has received funding from the European Union's $7^{th}$ Framework Programme, Marie Curie Actions-Initial Training Network, under grant agreement $n^o 607643$, ``Metric Analysis For Emergent Technologies (MAnET)". It was also supported by the H\'e Programme of Innovation, which is partly financed by the Netherlands Organization for Scientific research (NWO) under grant No.~$ 629.001.003$. Part of the funding (that is the development of the mathematical theory part) was received from the European Research Council under the European Community's $7^{th}$ Framework Programme (FP7/2007-2014)/ERC
grant agreement No.~$335555$. We would like to show our gratitude to Dr. Gonzalo Sanguinetti who provided insight and expertise that greatly assisted the research and improved the manuscript.
%Part of funding was also received from the European Research Council under the European Community's $7^{th}$ Framework Programme (FP7/2007-2014)/ERC grant agreement No.~$335555$.
\end{acknowledgements}

\bibliographystyle{spbasic}      % basic style, author-year citations
\bibliography{manuscript}   % name your BibTeX data base

\begin{thebibliography}{35}
\providecommand{\natexlab}[1]{#1}
\providecommand{\url}[1]{{#1}}
\providecommand{\urlprefix}{URL }
\expandafter\ifx\csname urlstyle\endcsname\relax
  \providecommand{\doi}[1]{DOI~\discretionary{}{}{}#1}\else
  \providecommand{\doi}{DOI~\discretionary{}{}{}\begingroup
  \urlstyle{rm}\Url}\fi
\providecommand{\eprint}[2][]{\url{#2}}

\bibitem[{Abbasi-Sureshjani et~al(2015)Abbasi-Sureshjani, Smit-Ockeloen, Zhang,
  and ter Haar~Romeny}]{abbasi2015biologically}
Abbasi-Sureshjani S, Smit-Ockeloen I, Zhang J, ter Haar~Romeny B (2015)
  Biologically-inspired supervised vasculature segmentation in {SLO} retinal
  fundus images. In: Image Analysis and Recognition, Lecture Notes in Computer
  Science, vol 9164, Springer, pp 325--334

\bibitem[{Abbasi-Sureshjani et~al(2016)Abbasi-Sureshjani, Favali, Citti, Sarti,
  and Romeny}]{abbasi2016cortically}
Abbasi-Sureshjani S, Favali M, Citti G, Sarti A, Romeny BM (2016)
  Cortically-inspired spectral clustering for connectivity analysis in retinal
  images: Curvature integration. arXiv preprint arXiv:160808049

\bibitem[{Agrachev et~al(2009)Agrachev, Boscain, Gauthier, and
  Rossi}]{agrachev2009intrinsic}
Agrachev A, Boscain U, Gauthier JP, Rossi F (2009) The intrinsic hypoelliptic
  laplacian and its heat kernel on unimodular {Lie} groups. J Funct Anal
  256(8):2621--2655

\bibitem[{August and Zucker(2000)}]{august2000curve}
August J, Zucker SW (2000) The curve indicator random field: Curve organization
  via edge correlation. In: Perceptual organization for artificial vision
  systems, Springer, pp 265--288

\bibitem[{August and Zucker(2003)}]{august2003sketches}
August J, Zucker SW (2003) Sketches with curvature: The curve indicator random
  field and {M}arkov processes. IEEE Trans Pattern Anal Mach Intell
  25(4):387--400

\bibitem[{Bekkers et~al(2014)Bekkers, Duits, Berendschot, and ter
  Haar~Romeny}]{Bekkers:2014aa}
Bekkers E, Duits R, Berendschot T, ter Haar~Romeny B (2014) A multi-orientation
  analysis approach to retinal vessel tracking. J Math Imag Vis 49(3):583--610

\bibitem[{Bosking et~al(1997)Bosking, Zhang, Schofield, and
  Fitzpatrick}]{bosking1997orientation}
Bosking WH, Zhang Y, Schofield B, Fitzpatrick D (1997) Orientation selectivity
  and the arrangement of horizontal connections in tree shrew striate cortex. J
  Neurosci 17(6):2112--2127

\bibitem[{Chapman et~al(2002)Chapman, Dell'Omo, Sartini, Witt, Hughes, Thom,
  and Pedrinelli}]{chapman2002peripheral}
Chapman N, Dell'Omo G, Sartini M, Witt N, Hughes A, Thom S, Pedrinelli R (2002)
  Peripheral vascular disease is associated with abnormal arteriolar diameter
  relationships at bifurcations in the human retina. Clin Sci 103(2):111--116

\bibitem[{Cheng et~al(2014)Cheng, De, Zhang, Lin, and Li}]{cheng2014tracing}
Cheng L, De J, Zhang X, Lin F, Li H (2014) Tracing retinal blood vessels by
  matrix-forest theorem of directed graphs. In: Medical Image Computing and
  Computer-Assisted Intervention--MICCAI 2014, Springer, pp 626--633

\bibitem[{Citti and Sarti(2006)}]{citti2006cortical}
Citti G, Sarti A (2006) A cortical based model of perceptual completion in the
  roto-translation space. J Math Imag Vis 24(3):307--326

\bibitem[{De et~al(2016)De, Cheng, Zhang, Lin, Li, Ong, Yu, Yu, and
  Ahmed}]{de2016graph}
De J, Cheng L, Zhang X, Lin F, Li H, Ong K, Yu W, Yu Y, Ahmed S (2016) A
  graph-theoretical approach for tracing filamentary structures in neuronal and
  retinal images. IEEE Trans Med Imag 35(1):257

\bibitem[{Duits and van Almsick(2008)}]{duits2008explicit}
Duits R, van Almsick M (2008) The explicit solutions of linear left-invariant
  second order stochastic evolution equations on the {2D} {E}uclidean motion
  group. Q Appl Math 66(1):27--68

\bibitem[{Duits and Franken(2009)}]{duits2009line}
Duits R, Franken E (2009) Line enhancement and completion via linear left
  invariant scale spaces on {SE(2)}. In: International Conference on Scale
  Space and Variational Methods in Computer Vision, Springer-Verlag, Lecture
  Notes in Computer Science, vol 5567, pp 795--807

\bibitem[{Duits et~al(2007)Duits, Felsberg, Granlund, and ter
  Haar~Romeny}]{duits2007image}
Duits R, Felsberg M, Granlund G, ter Haar~Romeny B (2007) Image analysis and
  reconstruction using a wavelet transform constructed from a reducible
  representation of the {E}uclidean motion group. Int J Comput Vis
  72(1):79--102

\bibitem[{Estrada et~al(2015)Estrada, Tomasi, Schmidler, and
  Farsiu}]{estrada2015tree}
Estrada R, Tomasi C, Schmidler SC, Farsiu S (2015) Tree topology estimation.
  IEEE Trans Pattern Anal Mach Intell 37(8):1688--1701

\bibitem[{Favali et~al(2016)Favali, Abbasi-Sureshjani, ter Haar~Romeny, and
  Sarti}]{favali2016analysis}
Favali M, Abbasi-Sureshjani S, ter Haar~Romeny B, Sarti A (2016) Analysis of
  vessel connectivities in retinal images by cortically inspired spectral
  clustering. J Math Imag Vis 56(1):158--172

\bibitem[{Field et~al(1993)Field, Hayes, and Hess}]{field1993contour}
Field DJ, Hayes A, Hess RF (1993) Contour integration by the human visual
  system: Evidence for a local ``association field''. Vis Res 33(2):173--193

\bibitem[{Foracchia et~al(2005)Foracchia, Grisan, and Ruggeri}]{Foracchia2005}
Foracchia M, Grisan E, Ruggeri A (2005) Luminosity and contrast normalization
  in retinal images. Med Image Anal 9(3):179 -- 190

\bibitem[{Fraz et~al(2012)Fraz, Remagnino, Hoppe, Uyyanonvara, Rudnicka, Owen,
  and Barman}]{Fraz2012407}
Fraz M, Remagnino P, Hoppe A, Uyyanonvara B, Rudnicka A, Owen C, Barman S
  (2012) Blood vessel segmentation methodologies in retinal images -- a survey.
  Comput Meth Prog Bio 108(1):407 -- 433

\bibitem[{Geisler(2008)}]{geisler2008visual}
Geisler WS (2008) Visual perception and the statistical properties of natural
  scenes. Annu Rev Psychol 59:167--192

\bibitem[{Hu et~al(2015)Hu, Abr{\`a}moff, and Garvin}]{hu2015automated}
Hu Q, Abr{\`a}moff MD, Garvin MK (2015) Automated construction of arterial and
  venous trees in retinal images. J Med Imag 2(4):044,001

\bibitem[{Hubel and Wiesel(1962)}]{hubel1962receptive}
Hubel DH, Wiesel TN (1962) Receptive fields, binocular interaction and
  functional architecture in the cat's visual cortex. J Physiol 160(1):106

\bibitem[{Lam et~al(1992)Lam, Lee, and Suen}]{lam1992thinning}
Lam L, Lee SW, Suen CY (1992) Thinning methodologies-a comprehensive survey.
  {IEEE} Trans Pattern Anal Mach Intell 14(9):869--885

\bibitem[{Mumford(1994)}]{mumford1994elastica}
Mumford D (1994) Elastica and computer vision, Springer New York, chap
  Algebraic Geometry and its Applications, pp 491--506

\bibitem[{Perrinet and Bednar(2015)}]{perrinet2015edge}
Perrinet LU, Bednar JA (2015) Edge co-occurrences can account for rapid
  categorization of natural versus animal images. Sci Rep 5

\bibitem[{Robert and Casella(2005)}]{robert2013monte}
Robert CP, Casella G (2005) Monte Carlo Statistical Methods (Springer Texts in
  Statistics). Springer-Verlag New York, Inc.

\bibitem[{Sanguinetti et~al(2010)Sanguinetti, Citti, and
  Sarti}]{sanguinetti2010model}
Sanguinetti G, Citti G, Sarti A (2010) A model of natural image edge
  co-occurrence in the rototranslation group. J Vis 10(14):37

\bibitem[{Smith et~al(2004)Smith, Wang, Wong, Rochtchina, Klein, Leeder, and
  Mitchell}]{smith2004retinal}
Smith W, Wang JJ, Wong TY, Rochtchina E, Klein R, Leeder SR, Mitchell P (2004)
  Retinal arteriolar narrowing is associated with 5-year incident severe
  hypertension. {T}he {B}lue {M}ountains {E}ye {S}tudy. Hypertension
  44(4):442--447

\bibitem[{Staal et~al(2004)Staal, Abr{\`a}moff, Niemeijer, Viergever, and van
  Ginneken}]{staal2004}
Staal J, Abr{\`a}moff MD, Niemeijer M, Viergever MA, van Ginneken B (2004)
  Ridge-based vessel segmentation in color images of the retina. IEEE Trans Med
  Imag 23(4):501--509

\bibitem[{T{\"u}retken et~al(2012)T{\"u}retken, Benmansour, and
  Fua}]{turetken2012automated}
T{\"u}retken E, Benmansour F, Fua P (2012) Automated reconstruction of tree
  structures using path classifiers and mixed integer programming. In: IEEE
  Conference on Computer Vision and Pattern Recognition, pp 566--573

\bibitem[{Wertheimer(1938)}]{wertheimer1938laws}
Wertheimer M (1938) Laws of organization in perceptual forms. In: Ellis W (ed)
  A {S}ource {B}ook of {G}estalt {P}sychology, Routledge and Kegan Paul, pp
  71--88

\bibitem[{Williams and Jacobs(1997)}]{williams1997stochastic}
Williams LR, Jacobs DW (1997) Stochastic completion fields: A neural model of
  illusory contour shape and salience. Neural Comput 9(4):837--858

\bibitem[{Zelnik-Manor and Perona(2004)}]{zelnik2004self}
Zelnik-Manor L, Perona P (2004) Self-tuning spectral clustering. In: Adv.
  Neural Inform. Process. Syst., pp 1601--1608

\bibitem[{Zhang et~al(2016{\natexlab{a}})Zhang, Dashtbozorg, Bekkers, Pluim,
  Duits, and ter Haar~Romeny}]{zhang2016robust}
Zhang J, Dashtbozorg B, Bekkers E, Pluim J, Duits R, ter Haar~Romeny B
  (2016{\natexlab{a}}) Robust retinal vessel segmentation via locally adaptive
  derivative frames in orientation scores. IEEE Trans Med Imag PP:1--1

\bibitem[{Zhang et~al(2016{\natexlab{b}})Zhang, Duits, Sanguinetti, and ter
  Haar~Romeny}]{zhang2014numerical}
Zhang J, Duits R, Sanguinetti G, ter Haar~Romeny BM (2016{\natexlab{b}})
  Numerical approaches for linear left-invariant diffusions on {SE(2)}, their
  comparison to exact solutions, and their applications in retinal imaging.
  Numer Math J Chin Univ 9:1--50

\end{thebibliography}
\end{document}